\documentclass{article}

\usepackage{graphicx}
\usepackage{booktabs}
\usepackage{caption}
\usepackage{subcaption}
\usepackage{longtable}
\usepackage{multirow}
\usepackage[table]{xcolor}

\usepackage{amsfonts} 
\usepackage{amsmath} 
\usepackage{amssymb} 

\usepackage{xspace}
\PassOptionsToPackage{sort}{natbib}

\newcommand{\numholdout}{24\xspace} 
\newcommand{\holdoutsuccess}{44\%\xspace} 
\newcommand{\numexpertdemo}{11,108\xspace}
\newcommand{\numdaggerdemo}{14,769\xspace}
\newcommand{\numrobotdemos}{25,877\xspace}
\newcommand{\numrobots}{12\xspace}
\newcommand{\numoperators}{7\xspace}
\newcommand{\numdemohours}{125\xspace}
\newcommand{\numvideos}{18,726\xspace}

\newcommand{\eighttask}{\textsuperscript{8}\xspace}
\newcommand{\methodname}{BC-Z\xspace}
\newcommand{\expnumber}[2]{{#1}\mathrm{e}{#2}}
\newcommand{\D}{\mathcal{D}}
\newcommand{\loss}{\mathcal{L}}
\usepackage{wrapfig}
\usepackage{todonotes}
\newcommand*{\rebuttal}{\textcolor{black}}
\usepackage{hyperref}

\usepackage[ruled,vlined]{algorithm2e}
\SetKwInput{KwInput}{Input}

\usepackage[final]{corl_2021} 

\usepackage{titlesec}
\titlespacing\section{0pt}{0pt plus 2pt minus 2pt}{0pt plus 2pt minus 2pt}
\titlespacing\subsection{0pt}{3pt plus 4pt minus 2pt}{0pt plus 2pt minus 2pt}

\title{\methodname: Zero-Shot Task Generalization with Robotic Imitation Learning}

\author{Eric Jang$^{1*}$, Alex Irpan$^{1*}$, Mohi Khansari$^{2}$, Daniel Kappler$^{2}$, Frederik Ebert$^{3\dagger}$,\\ 
\textbf{Corey Lynch$^{1}$, Sergey Levine$^{1, 3}$, Chelsea Finn$^{1, 4}$}
\AND
\normalfont{$^1$Robotics at Google ~ $^2$Everyday Robots ~ $^3$UC Berkeley ~ $^4$Stanford University} \\ [1em]
\url{https://sites.google.com/view/bc-z/home}
\vspace{-1em}
}

\begin{document}

\renewcommand{\thefootnote}{\fnsymbol{footnote}}
\footnotetext[1]{Equal Contribution}
\footnotetext[2]{Work done while author was at Google}
\renewcommand{\thefootnote}{\arabic{footnote}}

\maketitle


    \begin{abstract}
In this paper, we study the problem of enabling a vision-based robotic manipulation system to generalize to novel tasks, a long-standing challenge in robot learning. We approach the challenge from an imitation learning perspective, aiming to study how scaling and broadening the data collected can facilitate such generalization. To that end, we develop an interactive and flexible imitation learning system that can learn from both demonstrations and interventions and can be conditioned on different forms of information that convey the task, including pre-trained embeddings of natural language or videos of humans performing the task. When scaling data collection on a real robot to more than 100 distinct tasks, we find that this system can perform \numholdout \emph{unseen} manipulation tasks with an average success rate of \holdoutsuccess, without any robot demonstrations for those tasks.
\end{abstract}



\section{Introduction}
    One of the grand challenges in robotics is to create a general-purpose robot capable of performing a multitude of tasks in unstructured environments based on arbitrary user commands.
    The key challenge in this endeavour is \emph{generalization}: the robot must handle new environments, recognize and manipulate objects it has not seen before, and understand the intent of a command it has never been asked to execute. 
    End-to-end learning from pixels is a flexible choice for modeling the behavior of such generalist robots,
    as it has minimal assumptions about the state representation of the world. With sufficient real-world data, these methods should in principle enable robots to generalize across new tasks, objects, and scenes without requiring hand-coded, task-specific representations. 
    However, realizing this goal has generally remained elusive.
    In this paper, we study the problem of enabling a robot to generalize zero-shot \rebuttal{or few-shot} to new vision-based manipulation tasks.

    We study this problem using the framework of imitation learning. Prior works on imitation learning have shown one-shot or zero-shot generalization to new objects~\cite{finn2017one,james2018task,yu2018one,bonardi2020learning,young2020visual} and to new object goal configurations~\cite{duan2017one,dasari2020transformers}. However, zero-shot generalization to new tasks remains a challenge, particularly when considering vision-based manipulation tasks that cover a breadth of skills (e.g., wiping, pushing, pick-and-place) with diverse objects. Achieving such generalization depends on solving challenges relating to scaling up data collection and learning algorithms for diverse data.
    
    \begin{figure}[h]
  \centering
  \vspace{-0.1cm}
    \includegraphics[width=\textwidth]{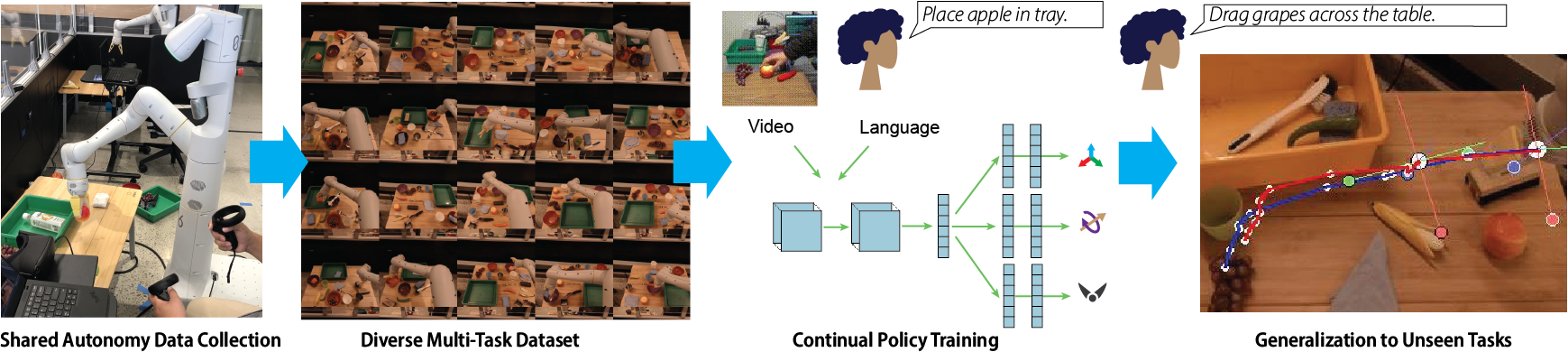} 
    \caption{\small Overview of \methodname. We collect a large-scale dataset (\numrobotdemos episodes) of  100 diverse manipulation tasks, and train a 7-DoF multi-task policy that conditions on task language strings or human video. We show this system produces a policy that is capable of generalizing zero-shot to new unseen tasks.}
  \label{fig:overview}
 \vspace{-10pt}
\end{figure}

    We develop an interactive imitation learning system with two key properties that enable high-quality data collection and generalization to entirely new tasks. First, our system incorporates shared autonomy into teleoperation to allow us to collect both raw demonstration data and human interventions to correct the robot's current policy. Second, our system flexibly conditions the policy on different forms of task specification, including a language instruction or a video of a person performing the task. Unlike discrete one-hot task identifiers~\cite{rahmatizadeh2018vision}, these continuous forms of task specification can in principle enable the robot to generalize zero-shot \rebuttal{or few-shot} to new tasks by providing a language or video command of the new task at test time.
    These properties have been explored previously; our aim is to empirically study whether these ideas scale to a broad range of real-world tasks.

    \rebuttal{Our main contribution is an empirical study of a large-scale interactive imitation learning system that solves a breadth of tasks, including zero-shot and few-shot generalization to tasks \emph{not seen} during training.}
    Using this system, we collect a large dataset of 100 robotic manipulation tasks, through a combination of expert teleoperation and a shared autonomy process where the human operator ``coaches'' the learned policy by fixing its mistakes.
    Across \numrobots robots, \numoperators different operators collected \numrobotdemos robot demonstrations that totaled \numdemohours hours of robot time, as well as \numvideos human videos of the same tasks. 
    At test time,
    the system is capable of performing \numholdout~unseen manipulation tasks between objects that have never previously appeared together in the same scene.
    These closed-loop visuomotor policies perform asynchronous inference and control at 10Hz, amounting to well over 100 decisions per episode. We open-source the demonstrations used to train this policy at \url{https://www.kaggle.com/google/bc-z-robot}.
\section{Related Work}
\label{sec:relatedwork}
Imitation learning has been successful in learning grasping and pick-place tasks from low-dimensional state~\cite{argall2009survey,khansariTRO2011,billard2004discovering,schaal2005learning,chalodhorn2007learning,pastor2009learning,mulling2013learning}. Deep learning has enabled imitation learning directly from raw image observations~\cite{pomerleau1989alvinn,zhang2018deep, rahmatizadeh2018vision}. In this work, we focus on enabling zero-shot \rebuttal{and few-shot} generalization to new tasks in an imitation learning framework.

Multiple prior imitation learning works have achieved different forms of generalization, including one-shot generalization to novel objects~\cite{finn2017one,james2018task,yu2018one,bonardi2020learning,zhou2019watch}, to novel object configurations~\cite{paine2018one}, and to novel goal configurations~\cite{duan2017one,huang2019neural,dasari2020transformers}, as well as zero-shot generalization to new objects~\cite{young2020visual}, scenes~\cite{pathak2018zero}, and goal configurations~\cite{goyal2021zero}. Many of these works adapt to the new scenario by conditioning on a robot demonstration~\cite{finn2017one,james2018task}, a video of a human~\cite{yu2018one,bonardi2020learning}, a language instruction~\cite{stepputtis2020language, lynch2020grounding}, or a goal image~\cite{pathak2018zero}.
Our system flexibly conditions on either a video of a human or a language instruction, and we focus on achieving zero-shot \rebuttal{(language) and few-shot (video)} generalization to \emph{entirely new} 7-DoF manipulation tasks on a real robot, including scenarios without goal images and where task-relevant objects are never encountered together in the training data. 

It is standard to collect demonstrations via teleoperation~\cite{calinon2009learning} or kinesthetic teaching~\cite{khansariTRO2011}, and active learning methods such as DAgger~\cite{ross2011reduction} help reduce distribution shift for the learner. Unfortunately, DAgger and some of its variants~\cite{ross2014aggrevate,laskey2017dart} are notoriously difficult to apply to robotic manipulation because they necessitate an interface where the expert must annotate the correct action when not in control of the robot policy. 
Inspired by recent works in autonomous driving, HG-DAgger~\cite{kelly2019hgdagger} and EIL~\cite{spencer2020eil}, our system instead only requires the expert to intervene when they believe the policy is likely to make an error and allows the expert to temporarily take full control to put the policy back on track. The resulting data collection scheme is easy to use and helps address distribution shift. Furthermore, the rate of expert interventions during data collection can be used as a live evaluation metric, which we empirically find correlates with policy success.

Beyond imitation learning, generalization has been studied in a number of other robot learning works. This includes works that generalize skills to novel objects~\cite{action_image,pinto2016supersizing,mahler2017dex,kalashnikov2018scalable,hatori2018interactively}, to novel environments~\cite{gupta2018robot}, from simulation to reality~\cite{sadeghi2018sim2real,tobin2018domain,mehta2020active,james2019sim,zhang2019vr}, and to new manipulation skills and objects~\cite{finn2017deep,dasari2019robonet,chebotar2021actionable,kalashnikov2021mt}. We focus on the last case of generalizing to novel tasks, but unlike these prior works, we tackle a large suite of 100 challenging tasks that involve 7 DoF control at 10 Hz and involve more than 100 decisions within an episode to solve the task.

\section{Problem Setup and Method Overview}

\begin{figure}[t]
\centering
\vspace{-0.2cm}
\includegraphics[width=\linewidth]{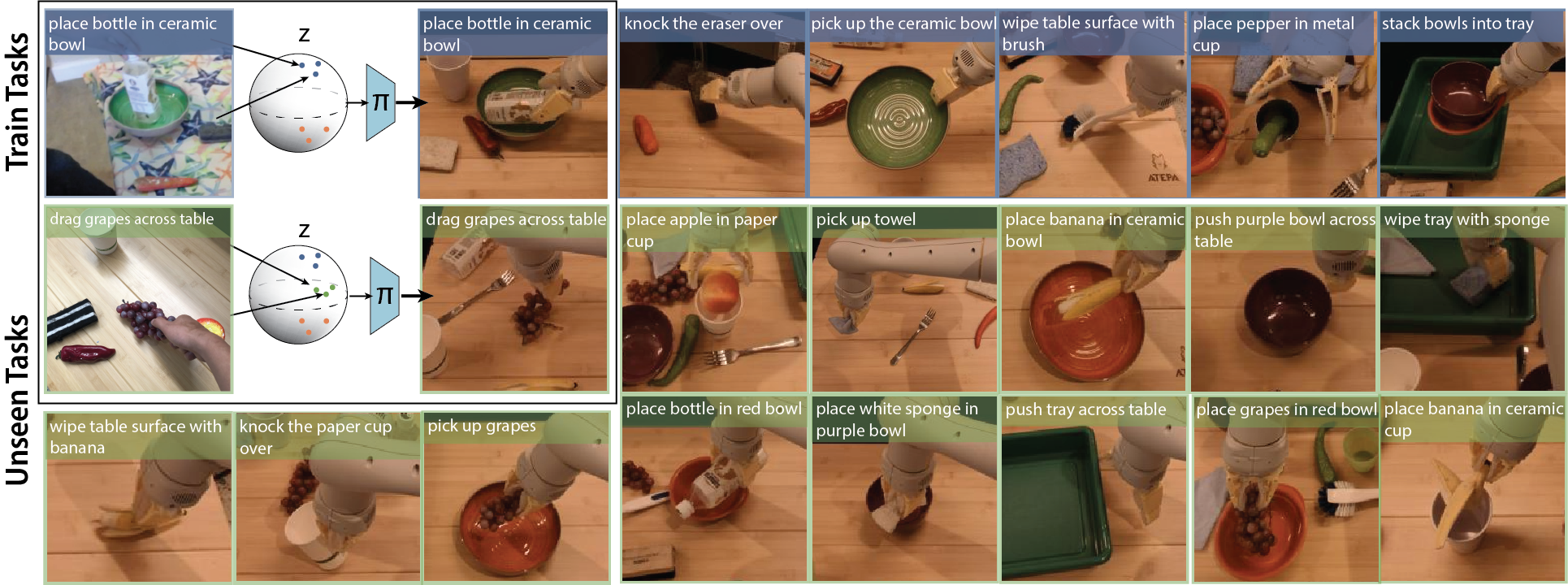} 
\caption{\small A subset of training tasks (top row), and a subset of held-out tasks (bottom two rows) used for evaluating zero shot task generalization. Top left: Given a pretrained task embedding computed from human videos or text, \methodname acts as an ``action decoder'' for the task embedding. 
}
\vspace{-0.3cm}
\label{fig:embed}
\end{figure}


An overview of our imitation learning system is shown in Figure~\ref{fig:overview}. Our goal is to train a conditional policy that can interpret RGB images, denoted $s \in \mathcal{S}$, together with a task command $w \in \mathcal{W}$,
which might correspond to a language string or a video of a person.
\rebuttal{Different tasks correspond to completing distinct objectives; some example tasks and corresponding commands are shown in Figure~\ref{fig:embed}.}
The policy is a mapping from images and commands to actions, and can be written as $\mu : \mathcal{S} \times \mathcal{W} \rightarrow \mathcal{A}$, where the action space $\mathcal{A}$ consists of the 6-DoF pose of the end effector as well as a 7$^\text{th}$ degree of freedom for continuous control of the parallel jaw gripper.

The policy is trained using a large-scale dataset collected via a VR-based teleoperation rig (see Figure~\ref{fig:overview}, left) through a combination of direct demonstration
and human-in-the-loop shared autonomy. In the latter,
trained policies are deployed on the robot, and the human operator intervenes to provide corrections when the robot makes a mistake. This procedure resembles the human-gated DAgger (HG-DAgger) algorithm~\cite{kelly2019hgdagger,ross2011reduction}, and provides iterative improvement for the learned policy, as well as a continuous signal that can be used to track the policy's performance. 

The policy architecture is divided into an encoder $q(z|w)$, which processes the command $w$ into an embedding $z \in \mathcal{Z}$,
and a control layer $\pi$, which processes $(s, z)$ to produce the action $a$, i.e. $\pi : \mathcal{S} \times \mathcal{Z} \rightarrow \mathcal{A}$. This decomposition is illustrated in Figure~\ref{fig:embed}, with further details in Section~\ref{sec:learning}. It provides our method with the ability to incorporate auxiliary supervision, such as pretrained language embeddings, which help to structure the latent task space and facilitate generalization. In our experiments, we will show that this enables generalization to tasks that were not seen during training, including novel compositions of verbs and objects.

\section{Data Collection and Workflow}
\label{sec:method}

In order for an imitation learning system to generalize to new tasks with zero demonstrations of said task, we must be able to easily collect a diverse dataset, provide corrective feedback, and evaluate many tasks at scale. In this section, we discuss these components of our system.


\textbf{System Setup.}
\label{method:tasks-and-system}
Our teleoperation system uses an Oculus VR headset which is attached to the robot's onboard computer via USB cable and tracks two handheld controllers.
The teleoperator stands behind the robot and uses the controllers to operate the robot with a line-of-sight 3rd-person view. The robot responds to the operator's movement in a 10 Hz non-realtime control loop. The relatively fast closed-loop control allows the operator to demonstrate a wide range of tasks with ease and quickly intervene if the robot is about to enter an unsafe state during autonomous execution.
Further details on the user interface and data collection are in Appendices~\ref{appendix:questcontrol} and~\ref{appendix:datacollect}.

\textbf{Environment and Tasks.}
We place each robot in front of a table with anywhere from 6 to 15 household objects with randomized poses. 
We collect demonstrations and videos of humans for 100 pre-specified tasks (listed in Tables~\ref{table:traintasks} and~\ref{table:remainingtrainsubtasks}), 
\rebuttal{which span 9 underlying skills such as pushing and pick-and-place}. The model is then evaluated on 29 \emph{new} tasks using a new language description or video of that task. For the method to perform well on these held-out tasks, it must both correctly interpret the new task command and output actions that are consistent with that task.


\textbf{Shared Autonomy Data Collection.}
\label{method:daggercollect}
Data collection begins with an initial expert-only phase, where
the human provides the demonstration of the task from start-to-finish.
After an initial multi-task policy is learned from expert-only data, we continue collecting demonstrations in ``shared autonomy'' mode, where the current policy attempts
the task while the human supervises. At any point the human may take over by gripping an ``override'' switch, which allows them to briefly take full control of the robot and perform necessary corrections when the policy is about to enter an unsafe state, or if they believe the current policy will not successfully complete the task.
This setup enables HG-DAgger~\cite{kelly2019hgdagger}, where intervention data is then aggregated with the existing data and used to re-train the policy. For the multi-task manipulation tasks, we collect \numexpertdemo expert-only demonstrations for the initial policies, then collected an additional \numdaggerdemo HG-DAgger demonstrations
covering 16 iterations of policy deployment, \rebuttal{where each iteration deploys the most recent policy trained on the aggregated dataset}. This gives a total of \numrobotdemos robot demos. We find in Table~\ref{table:ablation-results} that when controlling for the same number of total episodes, HG-DAgger improves performance substantially. 

\textbf{Shared Autonomy Evaluation.}
\label{method:daggereval}
When success rates are low, resources are best spent on collecting more data to improve the policy; but evaluation is also important to debug problems in the workflow. As the expected degree of generalization increases, we need more trials to evaluate the extent of policy generalization. This creates a resource trade-off: how should robot time be allocated between measuring policy success rates and collecting additional demonstrations to improve the policy? 
Fortunately, shared autonomy data collection confers an additional benefit: the \textit{intervention rate}, measured as the average number of interventions required per episode, can be used as an indication for policy performance. In Figure~\ref{fig:success-vs-autonomy}, we find that the intervention rate correlates negatively with overall policy success rate. 

\section{Learning Algorithm}
\label{sec:learning}

The data collection procedure above results in a large multi-task dataset. For each task $i$, this dataset contains expert data $(s,a) \in \mathcal{D}_e^i$, human video data $w_h \in \mathcal{D}_h^i$, and one language command $w_\ell^i$. We now discuss how we use this data to train the encoder $q(z|w)$ and the control layer $\pi(a|s, z)$.

\subsection{Language and Video Encoders}

Our encoder $q(z|w)$ takes either a language command $w_\ell^i$ or a video of a human $w_h$ as input and produces a task embedding $z$. If the command is a language command, we use a pretrained multilingual sentence encoder~\citep{yang2019multilingual}\footnote{Checkpoint from \url{https://tfhub.dev/google/universal-sentence-encoder-multilingual/3}} as our encoder, producing a 512-dim language vector for each task. Despite the simplicity, we find that these encoders work well in our experiments.

When task commands are instead a video of a human performing the task, we use a convolutional neural network to produce $z$, specifically a ResNet-18 based model.
Inspired by recent works~\cite{yu2018one,james2018task}, we train this network in an end-to-end manner. We collected a dataset of \numvideos videos of humans doing each training task, in a variety of home and office locations, camera viewpoints, and object configurations.
Using paired examples of a human video $w_h^i$ and corresponding demonstration demo $\{(s,a)\}^i$, we encode the human video $z^i\sim q(\cdot \mid w_h^i)$, then pass the embedding to the control layer $\pi(a|s,z^i)$, and then backpropagate gradient of the behavior cloning loss to both the policy and encoder parameters.

Visualizations of learned embeddings in Appendix~\ref{appendix:video} indicate that by itself, this end-to-end approach tends to overfit to initial object scenes, learn poor embeddings, and show poor task generalization. To help align the video embeddings more semantically, we therefore further introduce an auxiliary \textit{language regression} loss. Concretely, this auxiliary loss trains the video encoder to predict the embedding of the task's language command with a cosine loss. The resulting video encoder objective is as follows:
\begin{equation}
\min \sum_{\text{task } i} \sum_{\substack{
(s,a) \sim \mathcal{D}_e^i \\ w_h\sim\mathcal{D}_h^i \bigcup \mathcal{D}_e^i}} 
\underbrace{-\log \pi(a | s, z^i)}_{\text{behavior cloning}} + \underbrace{D_\text{cos}(z^i_h, z^i_\ell)}_{\text{language regression}} \text{, where } \underbrace{ z^i_h\sim q(\cdot | w_h)}_{\text{video encoder}}, \underbrace{ z^i_\ell \sim q(\cdot | w_\ell^i)}_{\text{language encoder}}
\end{equation}
where $D_\text{cos}$ denotes the cosine distance.
Since robot demos double as videos of the task, we also train encoded robot videos to match to the language vector. This language loss is critical to learning a more organized embedding space. Additional architecture and training details are in Appendix~\ref{appendix:video}.

\subsection{Policy Training}
\label{section:policy-training}

Given a fixed task embedding, we train $\pi(a|s,z)$ via Huber loss on XYZ and axis-angle predictions, and log loss for the gripper angle. During training, images are randomly cropped, downsampled, and subjected to standard photometric augmentations. Below we describe two additional design choices that we found to be helpful. Additional training details such as learning rates, batch sizes, pseudocode, and further hyperparameters are discussed in Appendix~\ref{appendix:training}.

\textbf{Open-Loop Auxiliary Predictions.} The policy predicts the action the robot would take, as well as an open-loop trajectory of the next 10 actions the policy would take if it were operating in an open-loop manner. At inference time, the policy operates closed-loop, only executing the first action based on the current image. The open-loop prediction confers an auxiliary training objective, and provides a way to visually inspect the quality of a closed-loop plan in an offline manner (see Figure~\ref{fig:overview}, right). 

\textbf{State Differences as Actions.} In standard imitation learning implementations, actions taken at demonstration-time are used directly as target labels to be predicted from states. However, cloning expert actions at 10Hz resulted in the policy learning very small actions, as well as dithering behavior. To address this, we define actions as state differences to target poses $N>1$ steps in the future, using an adaptive algorithm to choose $N$ based on how much the arm and gripper move. We provide ablation studies for this design choice in Section~\ref{results:ablationsingletask} and further details in Appendix~\ref{appendix:featurization}

\subsection{Network Architecture}
\label{sec:networkarchitecture}
\label{method:multitaskcond}
\begin{figure}[t]
  \centering
    \includegraphics[width=\linewidth]{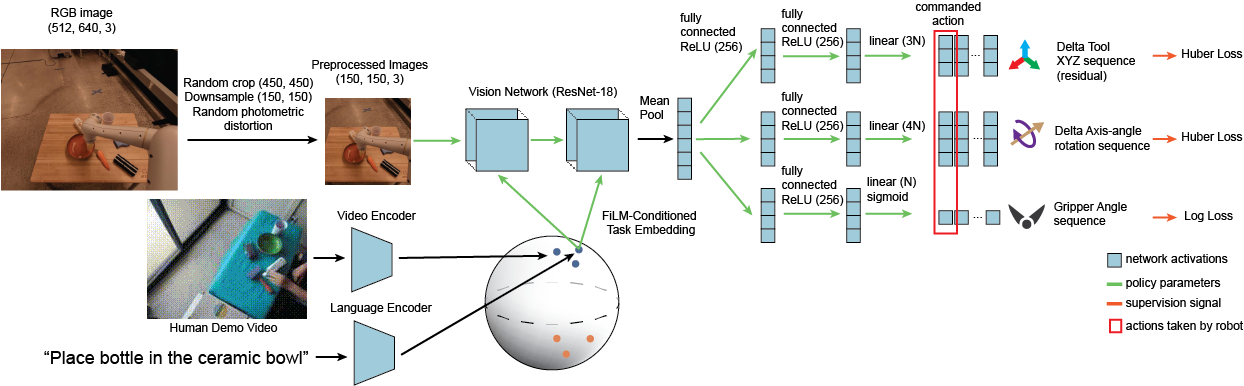} 
    \caption{\small \methodname network architecture. A monocular RGB image from the head-mounted camera is passed through a ResNet18 encoder, then through a two-layer MLP to predict each action modality (delta XYZ, delta axis-angle, and gripper angle). FiLM layers \cite{perez2018film} condition the architecture on a task embedding $z$ computed from language $w_\ell$ or video $w_h$. 
    } 
  \label{fig:architecture}
\end{figure}

We model the policy using a deep neural network, shown in Figure~\ref{fig:architecture}. 
The policy network processes the camera image with a ResNet18 ``torso'' \cite{he2016identity}, which branches from the last mean-pool layer into multiple ``action heads''. Each head is a multilayer perceptron with two hidden layers of size 256 each and ReLU activations, and models part of the end-effector action, specifically the delta XYZ, delta axis-angle, and normalized gripper angle. 
The policy is conditioned on a 512-dim task embedding $z$, through FiLM layers~\citep{perez2018film}. Following~\citet{perez2018film}, the task conditioning is linearly projected to channel-wise scales and shifts for each channel of each of the 4 ResNet blocks.

\section{Experimental Results}
\label{sec:results}

Our experiments aim to evaluate \methodname in large-scale imitation learning settings. We start with an initial validation of \methodname on single-task visual imitation learning. Then, our experiments will aim to answer the following questions: (1) Can \methodname enable zero-shot \rebuttal{and few-shot} generalization to new tasks from a command in the form of language or a video of a human? (2) Is the performance of \methodname bottlenecked by the task embedding or by the policy? (3) How important are different components of \methodname, including HG-DAgger data collection and adaptive state diffs? We present experiments aimed at these questions in this section.


\subsection{\methodname on Single-Task Imitation Learning}
\label{sec:dooropening}

We first aim to verify that \methodname can learn individual vision-based tasks before considering the more challenging multi-task setting. We choose two tasks: a \textit{bin-emptying} task where the robot must grasp objects from a bin and drop them into an adjacent bin, and a \textit{door opening} task where the robot must push open a door while avoiding collisions. Both tasks use the architecture in Figure~\ref{fig:architecture}, except that the door opening task involves predicting the forward and yaw velocity of the base instead of controlling the arm. The bin-emptying dataset has 2,759 demonstrations, while the door opening dataset has 12,000 demonstrations collected across 24 meeting rooms and 36,000 demonstrations across 5 meeting rooms in simulation. Further task and dataset details are in Appendix~\ref{appendix:door-results}.

\begin{wraptable}{r}{7cm}
  \vspace{-12pt}
\small
 \caption{\small Single-task bin and door performance, average and standard deviation across runs.}
 \vspace{-0.1cm}
 \begin{tabular}{lll}
 \toprule
\textbf{Bin-Emptying} & \textbf{Picks / Minute} & \textbf{\# Runs}\\
\midrule
Human Expert & 6.3 (2.1) & 2759 \\
\methodname (2759 demos) & 3.4 (1.2) & 9 \\
\midrule
\textbf{Door Opening} & \textbf{Success Rate} & \textbf{\# Runs}\\
\midrule
\methodname (24 Train Doors) & 87\% (2.2) & 480 \\
\methodname (4 Holdout Doors) & 94\% (2.7) & 80 \\

\bottomrule
\end{tabular}
  \label{table:door-results}
  
  \vspace{-15pt}
\end{wraptable}

In Table~\ref{table:door-results}, we see that the \methodname model is able to reach a pick-rate of 3.4 picks per minute,
over half the speed of a human teleoperator.
Further, we see that \methodname reaches a success rate of $87\%$ on the training door scenes and $94\%$ on held-out door scenes. These results validate that the \methodname model and data collect system can achieve good performance on both training and held-out scenes in the single-task setting. Additional analysis is provided in Appendix~\ref{appendix:sorty-toy-results}.
\subsection{Evaluating Zero-Shot \rebuttal{and Few-Shot} Task Generalization}
\label{sec:metatidy104}

\begin{figure}[t]
  \centering
    \includegraphics[width=\linewidth]{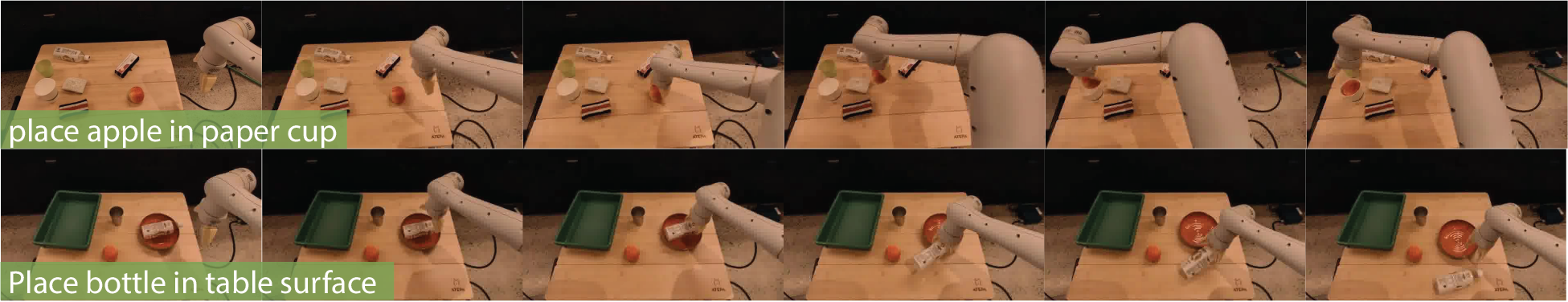}
    \caption{\small Qualitative examples of \methodname successfully performing held-out tasks.}
    \label{fig:holdoutqualitative}
    \vspace{-10pt}
\end{figure}

Next, we aim to test whether \methodname can achieve generalization to new tasks. Demonstrations are collected across 100 different manipulation tasks, comprising two disjoint sets of objects. Using disjoint sets of objects allows us to specifically test generalization to combinations of object-object pairs and object-action pairs that are not seen together during training.
For the first set of objects, demonstrations are collected across 21 different tasks, listed in Table~\ref{table:traintasks}, which cover a wide range of skills, from pick-and-place tasks to skills that require positioning the object in a certain way, like ``stand the bottle upright''. For the second set of objects, demonstrations are collected for 79 different tasks, including pick-and-place, surface wiping, and object stacking. The latter family has a smaller variety of manipulation behaviors, but is defined over a larger object set with more clutter. Object sets are shown in Appendix~\ref{appendix:datacollect} and a full list of train task sentences are in Appendix~\ref{appendix:subtask-results}.


We evaluate \methodname on 29 held-out tasks. \rebuttal{Language conditioned policies are given a novel sentence, while video conditioned policies are given the average embedding of a few human videos of the new task.} Four held-out tasks use objects in the 79-task family, whereas 25 tasks are generated by mixing objects between the 21-task family and 79-task family. Thus, the first 4 held-out tasks do not require cross-object set generalization, so they are easier to generalize to. Even so, we find that each of these 4 tasks are sufficiently challenging that training single-task policies on 300+ held-out demos with DAgger interventions completely fails, achieving 0\% task success. This provides a degree of calibration on the difficulty of these tasks. We hypothesize that a major contributing factor to this challenge is the wide range of locations, objects, and distractors that the skills must generalize to in our settings, as well as the wide range of these factors in the training data.



In Table~\ref{table:holdouttasks}, we see that language-conditioned \methodname is able to generalize zero-shot to both kinds of held-out tasks, averaging at 32\% success and showing non-zero success on \numholdout held-out tasks. Among the \numholdout hold-out tasks with non-zero success rates, \methodname achieves an average success of \holdoutsuccess when conditioned on language embeddings it has never seen. When conditioning on videos of humans, we find that generalization is much more difficult, but that \methodname is still able to generalize to nine novel tasks with a non-zero success rate, particularly when the task does not involve novel object combinations. 
Qualitatively, we observe that the language-conditioned policy usually moves towards the correct objects, clearly indicating that the task embedding is reflective of the correct task, as we further illustrate in the supplementary video. The most common source of failures are ``last-centimeter'' errors: failing to close the gripper, failing to let go of objects, or a near miss of the target object when letting go of an object in the gripper.

\begin{table}[t]
\scriptsize
  \caption{ Success rates for zero-shot \rebuttal{(language) and few-shot (video)} generalization to tasks not in the training dataset. The first 4 tasks only use objects from the 79-task family. The remaining tasks mix objects between the 21-task and 79-task families, requiring further generalization. Numbers in parentheses are 1 unit standard deviation. The language conditioning generalizes to several holdout tasks, whereas the video conditioning shows promise on tasks that do not mix objects between task families. Overall performance improves slightly with fewer distractor objects.
  }
  \label{table:holdouttasks}
\begin{center}
\begin{tabular}{p{1.2cm} p{4cm} p{2.2cm} p{2.2cm} p{2.2cm}}
\toprule
\textbf{Skill} &\textbf{Held-out tasks \newline (no demos during training)} &
\textbf{Lang-conditioned\newline (1 distractor)} & \textbf{Lang-conditioned\newline (4-5 distractors)} & \textbf{Video-conditioned\newline (4-5 distractors)} \\
\midrule
\multirow{3}{*}{pick-place} &`place sponge in tray' & 83\% (6.8) & 82\% (9.2) & 22\% (2.2) \\
&`place grapes in red bowl' & 87\% (6.2) & 75\% (10.8) & 12\% (7.8) \\
&`place apple in paper cup' & 30\% (8.4) & 33\% (12.2) & 14\% (7.8) \\
\arrayrulecolor{black!30}\cmidrule{2-5}
\multirow{1}{*}{pick-wipe} &`wipe tray with sponge' & 40\% (8.9) & 0\% (0) & 28\% (10.6) \\
\arrayrulecolor{black}\midrule
\multirow{11}{*}{pick-place} &`place banana in ceramic bowl' & 50\% (15.8) & 75\% (9.7) & 7.5\% (4.2) \\
&`place bottle in red bowl' & 50\% (15.8)& 75\% (9.7) & 0\% (0) \\
&`place grapes in ceramic bowl' & 70\% (14.5) & 70\% (10.3) & 0\% (0) \\
&`place bottle in table surface' & 0 & 50\% (11.2) & 5\% (3.5) \\
&`place white sponge in purple bowl' & 70\% (14.9)& 45\% (11.2) & 0\% (0) \\
&`place white sponge in tray' & 50\% (15.8) & 40\% (11.0) & 0\% (0) \\
&`place apple in ceramic bowl' & 30\% (14.5) & 20\% (8.9) & 0\% (0) \\
&`place bottle in purple bowl' & 30\% (14.5) & 20\% (8.9) & 0\% (0) \\
&`place banana in ceramic cup' & 10\% (9.5) & 0\% (0) & 0\% (0) \\
&`place banana on white sponge' & 40\% (15.5) & 0\% (0) & 0\% (0) \\
&`place metal cup in red bowl' & 0\% (0) & 0\% (0) & 0\% (0) \\
\arrayrulecolor{black!30}\cmidrule{2-5}
\multirow{6}{*}{grasp} &`pick up grapes' & 70\% (14.5) & 65\% (10.7) & 0\% (0) \\
&`pick up apple' & 20\% (12.7) & 55\% (11.2) & 5\% (3.5) \\
&`pick up towel' & 50\% (15.8) & 42.8\% (18.7) & 0\% (0) \\
&`pick up pepper' & 50\% (15.8)  & 35\% (10.7) & 12.5\% (5.2) \\
&`pick up bottle' & 40\% (15.5) & 30\% (10.3)& 17.5\% (6.0) \\
&`pick up the red bowl' & 30\% (14.5) & 0\% (0) & 0\% (0) \\
\arrayrulecolor{black!30}\cmidrule{2-5}
\multirow{1}{*}{pick-drag} &`drag grapes across the table' & 0\% (0) & 14\% (13.2) & 0\% (0) \\
\arrayrulecolor{black!30}\cmidrule{2-5}
\multirow{3}{*}{pick-wipe} &`wipe table surface with banana' & 0\% (0) & 10\% (6.7) & 0\% (0) \\
&`wipe tray with white sponge' & 20\% (12.7) & 0\% (0) & 0\% (0) \\
&`wipe ceramic bowl with brush' & 10\% (9.49) & 0\% (0) & 0\% (0) \\
\arrayrulecolor{black!30}\cmidrule{2-5}
\multirow{3}{*}{push} &`push purple bowl across the table' & 50\% (15.8) & 30\% (10.3) & 0\% (0) \\
&`push tray across the table' & 30\% (14.5) & 25\% (9.7) & 0\% (0) \\
&`push red bowl across the table' & 60\% (15.5) & 0\% (0) & 0\% (0) \\
\arrayrulecolor{black}\midrule
&\textit{Holdout Task Overall} & 38\% & 32\% & 4\% \\
\bottomrule
\end{tabular}
\end{center}
\vspace{-15pt}
\end{table}

\textbf{Is Performance Bottlenecked on the Encoder or the Policy?} Now that we see that \methodname can generalize to a substantial number of held-out tasks to some degree, we ask whether the performance is
\begin{wraptable}{r}{5.5cm}
\small
\caption{\small Training vs. generalization performance, averaged across 21 of the training tasks and all 28 held-out tasks.}
\vspace{-0.4cm}
\label{tbl:traintest}
\begin{tabular}{llc}\\\toprule  
\textbf{Setting} & \!\!\!\textbf{Task Conditioning}\!\!\! & \textbf{Success} \\
\midrule
\multirow{3}{*}{Train} & One-hot & 42\%\\  
& Language & 40\%\\  
& Video & 24\%\\  
\midrule
\multirow{2}{*}{Held-Out}& Language & 32\% \\
& Video & 4\% \\
\bottomrule
\end{tabular}\vspace{-0.4cm}
\end{wraptable}   limited more by the generalization of the encoder $q(z|w)$, the control layer $\pi(a|s,z)$, or both. To disentangle these factors, we measure the policy success rate on the training tasks conditioned in 
three ways:  a one-hot task identifier, language embeddings of the training task commands, and video embeddings of \textit{held-out} human videos of the training tasks. This comparison is in Table~\ref{tbl:traintest}.
The similar performance between one-hot and language suggests the latent language space is sufficient, and that language-conditioned performance on held-out tasks is bottlenecked on the control layer more than the embedding.
The more significant drop in performance of video-conditioned policies suggests inferring tasks from videos is much more difficult, particularly for held-out tasks.


\subsection{Ablation Studies and Comparisons}
\label{results:ablationsingletask}

We validate the importance of several \methodname design decisions using the (training) 21-task family. 
Our first set of ablations evaluate on the ``place the bottle in ceramic bowl'' command, which has the most demos (1000) of any task. We first test whether multi-task training is helpful for performance: we compare the multi-task system trained on \numrobotdemos demos across all tasks, to a single-task policy trained on just the 1000 demos for the target task. 
In Table~\ref{table:ablation-results} (left), the single-task baseline achieves just 5\% success. The low number is consistent with the low single-task performance on holdout tasks from Section~\ref{sec:metatidy104}: collecting data over several robots and operators likely makes the task harder to learn. Only when pooling data across many tasks does \methodname learn to solve the task. We ablate the adaptive state diff scheme described in Section~\ref{sec:networkarchitecture} and find that it is important; when naively choosing the $N=1$ future state to compute the expert actions, the policy fits the noise and moves too slowly, resulting in state drift away from good trajectories. 

\begin{table}[t]
\vspace{-15pt}
\small
  \caption{\small Ablation Studies. Left: Multi-task vs. single task models on the `place the bottle in the ceramic bowl' task. Training across tasks and with adaptive state-diffs is important for good training performance. Right: DAgger comparison on 'place the bottle in the ceramic bowl' (1-Task) and the 8-Task subset from Table~\ref{table:traintasks}. Controlled for the same amount of data, DAgger reaches higher success numbers significantly more quickly.}
  \label{table:ablation-results}
\begin{center}
\begin{tabular}{llll}
\toprule
\textbf{Method} & \textbf{1-Task} \\
\midrule
Multi-task, language conditioned & 52\% (6.3) \\
Multi-task, one-hot conditioned & 45\% (5.3) \\ 
Single-task baseline (1000 demos) & 5\% (2.8) \\ 
Multi-task, one-hot, no adaptive state-diff & 3\% (2.3) \\ 
\bottomrule
\end{tabular}
\hfill
\begin{tabular}{p{2.5cm} p{1.3cm} p{1.3cm}}
\toprule
\textbf{Dataset} & \textbf{1-Task} & \textbf{8-Task}  \\
\midrule
100\% Manual & 27\% (5.2)  & 23\% (4.2)\\
50\% Manual +\newline 50\% HG-DAgger  & 53\% (5.8) &  47\% (5.2)\\
\bottomrule
\end{tabular}
\end{center}
  \vspace{-20pt}
\end{table}

\begin{wrapfigure}{r}{0.44\textwidth}
\vspace{-20pt}
  \begin{center}
    \includegraphics[width=0.4\textwidth]{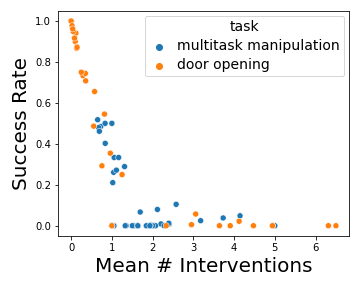}
  \end{center}
  \vspace{-0.4cm}
  \caption{\small Mean number of interventions vs. task success rate. Each point represents a policy evaluated during HG-DAgger data collection. There is a clear correlation between the mean number of interventions and success rate, suggesting that interventions can be used as a live proxy for performance.}
  \label{fig:success-vs-autonomy}
  \vspace{-20pt}
\end{wrapfigure}
We next ablate the use of HG-DAgger while keeping the total amount of data fixed. Specifically, we compare performance of policies trained using 50\% expert demos and 50\% HG-DAgger interventions, versus using 100\% expert demos. In Table~\ref{table:ablation-results} (right), we find that HG-DAgger significantly improves task performance over cloning expert demonstrations on both the `place bottle in ceramic bowl' task and 7 other training tasks. Further details on this comparison are in Appendix~\ref{app:ablationdetails}.
Finally, in Figure~\ref{fig:success-vs-autonomy}, we evaluate whether measuring HG-DAgger interventions can give us a live proxy of policy performance. We see that intervention frequency is inversely correlated with policy success, as measured by the fraction of successful episodes not requiring intervention. This result suggests that we can indeed use this metric with HG-DAgger for development purposes.



	


\section{Discussion}
\label{sec:conclusion}

We presented a multi-task imitation learning system that combines flexible task embeddings with large-scale training on a 100-task demonstration dataset, enabling it to generalize to entirely new tasks that were not seen in training based on user-provided language or video commands. Our evaluation covered 29 unseen vision-based manipulation tasks with a variety of objects and scenes. \rebuttal{The key conclusion of our empirical study is that simple imitation learning approaches can be scaled in a way that facilitates generalization to new tasks with zero additional robot data of those tasks. That is, we learn that we do not need more complex approaches to attain task-level generalization. Through the experiments, we also learn that 100 training tasks is sufficient for enabling generalization to new tasks, that HG-DAgger is important for good performance, and that frozen, pre-trained language embeddings make for excellent task conditioners without any additional training.}

Our system does have a number of limitations. First, the performance on novel tasks varies significantly. However, even for tasks that are less successful, the robot often exhibits behavior suggesting that it understands at least part of the task, reaching for the right object or performing a semantically related motion. This suggests that an exciting direction for future work is to use our policies as a general-purpose initialization for finetuning of downstream tasks, where additional training, perhaps with autonomous RL, could lead to significantly better performance. The structure of our language commands follows a simple ``(verb) (noun)'' structure. A direction to address this limitation is to relabel the dataset with a variety of human-provided annotations \cite{lynch2020grounding}, which could enable the system to handle more variability in the language structure. Another limitation is the lower performance of the video-conditioned policy, \rebuttal{which encourages future research on improving the generalization of video-based task representations and enhancing the performance of imitation learning algorithms as a whole, as low-level control errors are also a major bottleneck.}

\clearpage
\acknowledgments{Eric Jang, Alex Irpan, and Frederik Ebert ran experiments on different forms of task conditioning in the task generalization setup. Mohi Khansari helped build the HG-DAgger interface and ran experiments in the bin-emptying and door opening tasks. Daniel Kappler built the data annotation visualizer. Corey Lynch advised Frederik's internship and gave pointers on language models. Sergey Levine and Chelsea Finn supervised the project.

We would like to give special thanks to Noah Brown, Omar Cortes, Armando Fuentes, Kyle Jeffrey, Linda Luu, Sphurti Kirit More, Jornell Quiambao, Jarek Rettinghouse, Diego Reyes, Rosario Jauregui Ruano, and Clayton Tan for overseeing robot operations and collecting human videos of the tasks, as well as Jeffrey Bingham and Kanishka Rao for valuable discussions.}


\bibliography{references}  

\begin{thebibliography}{54}
\providecommand{\natexlab}[1]{#1}
\providecommand{\url}[1]{\texttt{#1}}
\expandafter\ifx\csname urlstyle\endcsname\relax
  \providecommand{\doi}[1]{doi: #1}\else
  \providecommand{\doi}{doi: \begingroup \urlstyle{rm}\Url}\fi

\bibitem[Finn et~al.(2017)Finn, Yu, Zhang, Abbeel, and Levine]{finn2017one}
C.~Finn, T.~Yu, T.~Zhang, P.~Abbeel, and S.~Levine.
\newblock One-shot visual imitation learning via meta-learning.
\newblock In \emph{Conference on Robot Learning}, pages 357--368. PMLR, 2017.

\bibitem[James et~al.(2018)James, Bloesch, and Davison]{james2018task}
S.~James, M.~Bloesch, and A.~J. Davison.
\newblock Task-embedded control networks for few-shot imitation learning.
\newblock In \emph{Conference on Robot Learning}, pages 783--795. PMLR, 2018.

\bibitem[Yu et~al.(2018)Yu, Finn, Xie, Dasari, Zhang, Abbeel, and
  Levine]{yu2018one}
T.~Yu, C.~Finn, A.~Xie, S.~Dasari, T.~Zhang, P.~Abbeel, and S.~Levine.
\newblock One-shot imitation from observing humans via domain-adaptive
  meta-learning.
\newblock \emph{Robotics: Science and Systems}, 2018.

\bibitem[Bonardi et~al.(2020)Bonardi, James, and Davison]{bonardi2020learning}
A.~Bonardi, S.~James, and A.~J. Davison.
\newblock Learning one-shot imitation from humans without humans.
\newblock \emph{IEEE Robotics and Automation Letters}, 5\penalty0 (2):\penalty0
  3533--3539, 2020.

\bibitem[Young et~al.(2020)Young, Gandhi, Tulsiani, Gupta, Abbeel, and
  Pinto]{young2020visual}
S.~Young, D.~Gandhi, S.~Tulsiani, A.~Gupta, P.~Abbeel, and L.~Pinto.
\newblock Visual imitation made easy.
\newblock \emph{arXiv e-prints}, pages arXiv--2008, 2020.

\bibitem[Duan et~al.(2017)Duan, Andrychowicz, Stadie, Ho, Schneider, Sutskever,
  Abbeel, and Zaremba]{duan2017one}
Y.~Duan, M.~Andrychowicz, B.~C. Stadie, J.~Ho, J.~Schneider, I.~Sutskever,
  P.~Abbeel, and W.~Zaremba.
\newblock One-shot imitation learning.
\newblock \emph{arXiv preprint arXiv:1703.07326}, 2017.

\bibitem[Dasari and Gupta(2020)]{dasari2020transformers}
S.~Dasari and A.~Gupta.
\newblock Transformers for one-shot visual imitation.
\newblock \emph{Conference on Robot Learning (CoRL)}, 2020.

\bibitem[Rahmatizadeh et~al.(2018)Rahmatizadeh, Abolghasemi, B{\"o}l{\"o}ni,
  and Levine]{rahmatizadeh2018vision}
R.~Rahmatizadeh, P.~Abolghasemi, L.~B{\"o}l{\"o}ni, and S.~Levine.
\newblock Vision-based multi-task manipulation for inexpensive robots using
  end-to-end learning from demonstration.
\newblock In \emph{2018 IEEE international conference on robotics and
  automation (ICRA)}, pages 3758--3765. IEEE, 2018.

\bibitem[Argall et~al.(2009)Argall, Chernova, Veloso, and
  Browning]{argall2009survey}
B.~D. Argall, S.~Chernova, M.~Veloso, and B.~Browning.
\newblock A survey of robot learning from demonstration.
\newblock \emph{Robotics and autonomous systems}, 57\penalty0 (5):\penalty0
  469--483, 2009.

\bibitem[Khansari-Zadeh and Billard(2011)]{khansariTRO2011}
S.~Khansari-Zadeh and A.~Billard.
\newblock Learning stable non-linear dynamical systems with gaussian mixture
  models.
\newblock \emph{IEEE Transaction on Robotics}, 27\penalty0 (5):\penalty0
  943--957, 2011.

\bibitem[Billard et~al.(2004)Billard, Epars, Calinon, Schaal, and
  Cheng]{billard2004discovering}
A.~Billard, Y.~Epars, S.~Calinon, S.~Schaal, and G.~Cheng.
\newblock Discovering optimal imitation strategies.
\newblock \emph{Robotics and autonomous systems}, 47\penalty0 (2-3):\penalty0
  69--77, 2004.

\bibitem[Schaal et~al.(2005)Schaal, Peters, Nakanishi, and
  Ijspeert]{schaal2005learning}
S.~Schaal, J.~Peters, J.~Nakanishi, and A.~Ijspeert.
\newblock Learning movement primitives.
\newblock In \emph{Robotics research. the eleventh international symposium},
  pages 561--572. Springer, 2005.

\bibitem[Chalodhorn et~al.(2007)Chalodhorn, Grimes, Grochow, and
  Rao]{chalodhorn2007learning}
R.~Chalodhorn, D.~B. Grimes, K.~Grochow, and R.~P. Rao.
\newblock Learning to walk through imitation.
\newblock In \emph{IJCAI}, volume~7, pages 2084--2090, 2007.

\bibitem[Pastor et~al.(2009)Pastor, Hoffmann, Asfour, and
  Schaal]{pastor2009learning}
P.~Pastor, H.~Hoffmann, T.~Asfour, and S.~Schaal.
\newblock Learning and generalization of motor skills by learning from
  demonstration.
\newblock In \emph{2009 IEEE International Conference on Robotics and
  Automation}, pages 763--768. IEEE, 2009.

\bibitem[M{\"u}lling et~al.(2013)M{\"u}lling, Kober, Kroemer, and
  Peters]{mulling2013learning}
K.~M{\"u}lling, J.~Kober, O.~Kroemer, and J.~Peters.
\newblock Learning to select and generalize striking movements in robot table
  tennis.
\newblock \emph{The International Journal of Robotics Research}, 32\penalty0
  (3):\penalty0 263--279, 2013.

\bibitem[Pomerleau(1989)]{pomerleau1989alvinn}
D.~A. Pomerleau.
\newblock Alvinn: An autonomous land vehicle in a neural network.
\newblock Technical report, Carnegie-Mellon University, 1989.

\bibitem[Zhang et~al.(2018)Zhang, McCarthy, Jow, Lee, Chen, Goldberg, and
  Abbeel]{zhang2018deep}
T.~Zhang, Z.~McCarthy, O.~Jow, D.~Lee, X.~Chen, K.~Goldberg, and P.~Abbeel.
\newblock Deep imitation learning for complex manipulation tasks from virtual
  reality teleoperation.
\newblock In \emph{2018 IEEE International Conference on Robotics and
  Automation (ICRA)}, pages 5628--5635. IEEE, 2018.

\bibitem[Zhou et~al.(2020)Zhou, Jang, Kappler, Herzog, Khansari, Wohlhart, Bai,
  Kalakrishnan, Levine, and Finn]{zhou2019watch}
A.~Zhou, E.~Jang, D.~Kappler, A.~Herzog, M.~Khansari, P.~Wohlhart, Y.~Bai,
  M.~Kalakrishnan, S.~Levine, and C.~Finn.
\newblock Watch, try, learn: Meta-learning from demonstrations and reward.
\newblock \emph{International Conference on Learning Representations (ICLR)},
  2020.

\bibitem[Paine et~al.(2018)Paine, Colmenarejo, Wang, Reed, Aytar, Pfaff,
  Hoffman, Barth-Maron, Cabi, Budden, et~al.]{paine2018one}
T.~L. Paine, S.~G. Colmenarejo, Z.~Wang, S.~Reed, Y.~Aytar, T.~Pfaff, M.~W.
  Hoffman, G.~Barth-Maron, S.~Cabi, D.~Budden, et~al.
\newblock One-shot high-fidelity imitation: Training large-scale deep nets with
  rl.
\newblock \emph{arXiv preprint arXiv:1810.05017}, 2018.

\bibitem[Huang et~al.(2019)Huang, Nair, Xu, Zhu, Garg, Fei-Fei, Savarese, and
  Niebles]{huang2019neural}
D.-A. Huang, S.~Nair, D.~Xu, Y.~Zhu, A.~Garg, L.~Fei-Fei, S.~Savarese, and
  J.~C. Niebles.
\newblock Neural task graphs: Generalizing to unseen tasks from a single video
  demonstration.
\newblock In \emph{Proceedings of the IEEE/CVF Conference on Computer Vision
  and Pattern Recognition}, pages 8565--8574, 2019.

\bibitem[Pathak et~al.(2018)Pathak, Mahmoudieh, Luo, Agrawal, Chen, Shentu,
  Shelhamer, Malik, Efros, and Darrell]{pathak2018zero}
D.~Pathak, P.~Mahmoudieh, G.~Luo, P.~Agrawal, D.~Chen, Y.~Shentu, E.~Shelhamer,
  J.~Malik, A.~A. Efros, and T.~Darrell.
\newblock Zero-shot visual imitation.
\newblock In \emph{Proceedings of the IEEE conference on computer vision and
  pattern recognition workshops}, pages 2050--2053, 2018.

\bibitem[Goyal et~al.(2021)Goyal, Mooney, and Niekum]{goyal2021zero}
P.~Goyal, R.~J. Mooney, and S.~Niekum.
\newblock Zero-shot task adaptation using natural language.
\newblock \emph{arXiv preprint arXiv:2106.02972}, 2021.

\bibitem[Stepputtis et~al.(2020)Stepputtis, Campbell, Phielipp, Lee, Baral, and
  Amor]{stepputtis2020language}
S.~Stepputtis, J.~Campbell, M.~Phielipp, S.~Lee, C.~Baral, and H.~B. Amor.
\newblock Language-conditioned imitation learning for robot manipulation tasks.
\newblock \emph{arXiv preprint arXiv:2010.12083}, 2020.

\bibitem[Lynch and Sermanet(2020)]{lynch2020grounding}
C.~Lynch and P.~Sermanet.
\newblock Grounding language in play.
\newblock \emph{arXiv preprint arXiv:2005.07648}, 2020.

\bibitem[Calinon et~al.(2009)Calinon, Evrard, Gribovskaya, Billard, and
  Kheddar]{calinon2009learning}
S.~Calinon, P.~Evrard, E.~Gribovskaya, A.~Billard, and A.~Kheddar.
\newblock Learning collaborative manipulation tasks by demonstration using a
  haptic interface.
\newblock In \emph{2009 International Conference on Advanced Robotics}, pages
  1--6. IEEE, 2009.

\bibitem[Ross et~al.(2011)Ross, Gordon, and Bagnell]{ross2011reduction}
S.~Ross, G.~Gordon, and D.~Bagnell.
\newblock A reduction of imitation learning and structured prediction to
  no-regret online learning.
\newblock In \emph{Proceedings of the fourteenth international conference on
  artificial intelligence and statistics}, pages 627--635. JMLR Workshop and
  Conference Proceedings, 2011.

\bibitem[Ross and Bagnell(2014)]{ross2014aggrevate}
S.~Ross and J.~A. Bagnell.
\newblock Reinforcement and imitation learning via interactive no-regret
  learning.
\newblock \emph{arXiv preprint arXiv:1406.5979}, 2014.

\bibitem[Laskey et~al.(2017)Laskey, Lee, Fox, Dragan, and
  Goldberg]{laskey2017dart}
M.~Laskey, J.~Lee, R.~Fox, A.~Dragan, and K.~Goldberg.
\newblock Dart: Noise injection for robust imitation learning.
\newblock In \emph{Conference on robot learning}, pages 143--156. PMLR, 2017.

\bibitem[Kelly et~al.(2019)Kelly, Sidrane, Driggs-Campbell, and
  Kochenderfer]{kelly2019hgdagger}
M.~Kelly, C.~Sidrane, K.~Driggs-Campbell, and M.~J. Kochenderfer.
\newblock Hg-dagger: Interactive imitation learning with human experts.
\newblock In \emph{2019 International Conference on Robotics and Automation
  (ICRA)}, pages 8077--8083. IEEE, 2019.

\bibitem[Spencer et~al.(2020)Spencer, Choudhury, Barnes, Schmittle, Chiang,
  Ramadge, and Srinivasa]{spencer2020eil}
J.~Spencer, S.~Choudhury, M.~Barnes, M.~Schmittle, M.~Chiang, P.~Ramadge, and
  S.~Srinivasa.
\newblock Learning from interventions: Human-robot interaction as both explicit
  and implicit feedback.
\newblock 07 2020.
\newblock \doi{10.15607/RSS.2020.XVI.055}.

\bibitem[Khansari et~al.(2020)Khansari, Kappler, Luo, Bingham, and
  Kalakrishnan]{action_image}
M.~Khansari, D.~Kappler, J.~Luo, J.~Bingham, and M.~Kalakrishnan.
\newblock Action image representation: Learning scalable deep grasping policies
  with zero real world data.
\newblock In \emph{2020 IEEE International Conference on Robotics and
  Automation (ICRA)}, pages 3597--3603, 2020.

\bibitem[Pinto and Gupta(2016)]{pinto2016supersizing}
L.~Pinto and A.~Gupta.
\newblock Supersizing self-supervision: Learning to grasp from 50k tries and
  700 robot hours.
\newblock In \emph{2016 IEEE international conference on robotics and
  automation (ICRA)}, pages 3406--3413. IEEE, 2016.

\bibitem[Mahler et~al.(2017)Mahler, Liang, Niyaz, Laskey, Doan, Liu, Ojea, and
  Goldberg]{mahler2017dex}
J.~Mahler, J.~Liang, S.~Niyaz, M.~Laskey, R.~Doan, X.~Liu, J.~A. Ojea, and
  K.~Goldberg.
\newblock Dex-net 2.0: Deep learning to plan robust grasps with synthetic point
  clouds and analytic grasp metrics.
\newblock \emph{arXiv preprint arXiv:1703.09312}, 2017.

\bibitem[Kalashnikov et~al.(2018)Kalashnikov, Irpan, Pastor, Ibarz, Herzog,
  Jang, Quillen, Holly, Kalakrishnan, Vanhoucke,
  et~al.]{kalashnikov2018scalable}
D.~Kalashnikov, A.~Irpan, P.~Pastor, J.~Ibarz, A.~Herzog, E.~Jang, D.~Quillen,
  E.~Holly, M.~Kalakrishnan, V.~Vanhoucke, et~al.
\newblock Scalable deep reinforcement learning for vision-based robotic
  manipulation.
\newblock In \emph{Conference on Robot Learning}, pages 651--673. PMLR, 2018.

\bibitem[Hatori et~al.(2018)Hatori, Kikuchi, Kobayashi, Takahashi, Tsuboi,
  Unno, Ko, and Tan]{hatori2018interactively}
J.~Hatori, Y.~Kikuchi, S.~Kobayashi, K.~Takahashi, Y.~Tsuboi, Y.~Unno, W.~Ko,
  and J.~Tan.
\newblock Interactively picking real-world objects with unconstrained spoken
  language instructions.
\newblock In \emph{IEEE International Conference on Robotics and Automation
  (ICRA)}, 2018.

\bibitem[Gupta et~al.(2018)Gupta, Murali, Gandhi, and Pinto]{gupta2018robot}
A.~Gupta, A.~Murali, D.~Gandhi, and L.~Pinto.
\newblock Robot learning in homes: Improving generalization and reducing
  dataset bias.
\newblock \emph{arXiv preprint arXiv:1807.07049}, 2018.

\bibitem[Sadeghi et~al.(2018)Sadeghi, Toshev, Jang, and
  Levine]{sadeghi2018sim2real}
F.~Sadeghi, A.~Toshev, E.~Jang, and S.~Levine.
\newblock Sim2real viewpoint invariant visual servoing by recurrent control.
\newblock In \emph{Proceedings of the IEEE Conference on Computer Vision and
  Pattern Recognition}, pages 4691--4699, 2018.

\bibitem[Tobin et~al.(2018)Tobin, Biewald, Duan, Andrychowicz, Handa, Kumar,
  McGrew, Ray, Schneider, Welinder, et~al.]{tobin2018domain}
J.~Tobin, L.~Biewald, R.~Duan, M.~Andrychowicz, A.~Handa, V.~Kumar, B.~McGrew,
  A.~Ray, J.~Schneider, P.~Welinder, et~al.
\newblock Domain randomization and generative models for robotic grasping.
\newblock In \emph{2018 IEEE/RSJ International Conference on Intelligent Robots
  and Systems (IROS)}, pages 3482--3489. IEEE, 2018.

\bibitem[Mehta et~al.(2020)Mehta, Diaz, Golemo, Pal, and
  Paull]{mehta2020active}
B.~Mehta, M.~Diaz, F.~Golemo, C.~J. Pal, and L.~Paull.
\newblock Active domain randomization.
\newblock In \emph{Conference on Robot Learning}, pages 1162--1176. PMLR, 2020.

\bibitem[James et~al.(2019)James, Wohlhart, Kalakrishnan, Kalashnikov, Irpan,
  Ibarz, Levine, Hadsell, and Bousmalis]{james2019sim}
S.~James, P.~Wohlhart, M.~Kalakrishnan, D.~Kalashnikov, A.~Irpan, J.~Ibarz,
  S.~Levine, R.~Hadsell, and K.~Bousmalis.
\newblock Sim-to-real via sim-to-sim: Data-efficient robotic grasping via
  randomized-to-canonical adaptation networks.
\newblock In \emph{Proceedings of the IEEE/CVF Conference on Computer Vision
  and Pattern Recognition}, pages 12627--12637, 2019.

\bibitem[Zhang et~al.(2019)Zhang, Tai, Yun, Xiong, Liu, Boedecker, and
  Burgard]{zhang2019vr}
J.~Zhang, L.~Tai, P.~Yun, Y.~Xiong, M.~Liu, J.~Boedecker, and W.~Burgard.
\newblock Vr-goggles for robots: Real-to-sim domain adaptation for visual
  control.
\newblock \emph{IEEE Robotics and Automation Letters}, 4\penalty0 (2):\penalty0
  1148--1155, 2019.

\bibitem[Finn and Levine(2017)]{finn2017deep}
C.~Finn and S.~Levine.
\newblock Deep visual foresight for planning robot motion.
\newblock In \emph{2017 IEEE International Conference on Robotics and
  Automation (ICRA)}, pages 2786--2793. IEEE, 2017.

\bibitem[Dasari et~al.(2019)Dasari, Ebert, Tian, Nair, Bucher, Schmeckpeper,
  Singh, Levine, and Finn]{dasari2019robonet}
S.~Dasari, F.~Ebert, S.~Tian, S.~Nair, B.~Bucher, K.~Schmeckpeper, S.~Singh,
  S.~Levine, and C.~Finn.
\newblock Robonet: Large-scale multi-robot learning.
\newblock \emph{arXiv preprint arXiv:1910.11215}, 2019.

\bibitem[Chebotar et~al.(2021)Chebotar, Hausman, Lu, Xiao, Kalashnikov, Varley,
  Irpan, Eysenbach, Julian, Finn, et~al.]{chebotar2021actionable}
Y.~Chebotar, K.~Hausman, Y.~Lu, T.~Xiao, D.~Kalashnikov, J.~Varley, A.~Irpan,
  B.~Eysenbach, R.~Julian, C.~Finn, et~al.
\newblock Actionable models: Unsupervised offline reinforcement learning of
  robotic skills.
\newblock \emph{arXiv preprint arXiv:2104.07749}, 2021.

\bibitem[Kalashnikov et~al.(2021)Kalashnikov, Varley, Chebotar, Swanson,
  Jonschkowski, Finn, Levine, and Hausman]{kalashnikov2021mt}
D.~Kalashnikov, J.~Varley, Y.~Chebotar, B.~Swanson, R.~Jonschkowski, C.~Finn,
  S.~Levine, and K.~Hausman.
\newblock Mt-opt: Continuous multi-task robotic reinforcement learning at
  scale.
\newblock \emph{arXiv preprint arXiv:2104.08212}, 2021.

\bibitem[Yang et~al.(2019)Yang, Cer, Ahmad, Guo, Law, Constant, Abrego, Yuan,
  Tar, Sung, et~al.]{yang2019multilingual}
Y.~Yang, D.~Cer, A.~Ahmad, M.~Guo, J.~Law, N.~Constant, G.~H. Abrego, S.~Yuan,
  C.~Tar, Y.-H. Sung, et~al.
\newblock Multilingual universal sentence encoder for semantic retrieval.
\newblock \emph{arXiv preprint arXiv:1907.04307}, 2019.

\bibitem[Perez et~al.(2018)Perez, Strub, De~Vries, Dumoulin, and
  Courville]{perez2018film}
E.~Perez, F.~Strub, H.~De~Vries, V.~Dumoulin, and A.~Courville.
\newblock Film: Visual reasoning with a general conditioning layer.
\newblock In \emph{Proceedings of the AAAI Conference on Artificial
  Intelligence}, volume~32, 2018.

\bibitem[He et~al.(2016)He, Zhang, Ren, and Sun]{he2016identity}
K.~He, X.~Zhang, S.~Ren, and J.~Sun.
\newblock Identity mappings in deep residual networks.
\newblock In \emph{European conference on computer vision}, pages 630--645.
  Springer, 2016.

\bibitem[Huber(1992)]{huber1992robust}
P.~J. Huber.
\newblock Robust estimation of a location parameter.
\newblock In \emph{Breakthroughs in statistics}, pages 492--518. Springer,
  1992.

\bibitem[Oord et~al.(2018)Oord, Li, and Vinyals]{oord2018representation}
A.~v.~d. Oord, Y.~Li, and O.~Vinyals.
\newblock Representation learning with contrastive predictive coding.
\newblock \emph{arXiv preprint arXiv:1807.03748}, 2018.

\bibitem[Ho et~al.(2021)Ho, Rao, Xu, Jang, Khansari, and Bai]{retinagan2021}
D.~Ho, K.~Rao, Z.~Xu, E.~Jang, M.~Khansari, and Y.~Bai.
\newblock Retinagan: An object-aware approach to sim-to-real transfer.
\newblock In \emph{2021 International Conference on Robotics and Automation
  (ICRA)}, 2021.

\bibitem[de~Haan et~al.(2019)de~Haan, Jayaraman, and Levine]{de2019causal}
P.~de~Haan, D.~Jayaraman, and S.~Levine.
\newblock Causal confusion in imitation learning.
\newblock \emph{arXiv preprint arXiv:1905.11979}, 2019.

\bibitem[Zhang et~al.(2018)Zhang, Cissé, Dauphin, and
  Lopez-Paz]{zhang2018mixup}
H.~Zhang, M.~Cissé, Y.~N. Dauphin, and D.~Lopez-Paz.
\newblock mixup: Beyond empirical risk minimization.
\newblock In \emph{ICLR (Poster)}, 2018.
\newblock URL \url{https://openreview.net/forum?id=r1Ddp1-Rb}.

\bibitem[Miech et~al.(2020)Miech, Alayrac, Smaira, Laptev, Sivic, and
  Zisserman]{miech2020end}
A.~Miech, J.-B. Alayrac, L.~Smaira, I.~Laptev, J.~Sivic, and A.~Zisserman.
\newblock End-to-end learning of visual representations from uncurated
  instructional videos.
\newblock In \emph{Proceedings of the IEEE/CVF Conference on Computer Vision
  and Pattern Recognition}, pages 9879--9889, 2020.

\end{thebibliography}

    \newpage
    
    \section*{Appendix}
    \appendix
    \setcounter{section}{0}
    
    \section{Teleoperation Interface}
    \label{appendix:questcontrol}
The human teleoperator holds two wireless Oculus Quest controllers and uses the same interface to perform demonstrations for all tasks. When the override button is held (the clutch on the right controller), the robot arm tracks the controller's position and orientation. The robot can be toggled between autonomous and manual mode. In manual mode, the robot stays still unless the operator moves it. In autonomous mode, the robot follows a learned policy, unless the operator overrides it.

\begin{table}[h]
 \caption{Teleoperation buttons and controls.}
\begin{tabular}{p{0.25\linewidth}|p{0.7\linewidth}}
\toprule
\textbf{Control} & \textbf{Function} \\
\midrule
\textit{Right Controller (Arm)} & \\

\midrule
A   & Start recording, or mark demo as success if already recording \\
B   & Stops current recording marking as failure (if applicable), then bring robot to reset pose \\
Clutch & Override policy and engage manual arm teleop until clutch is released \\
Trigger & Continuous gripper closure. Pressing the trigger all the way closes the gripper fully, and letting go of the trigger opens the gripper. \\
\midrule
\textit{Left Controller (Base)} & \\
\midrule
X   & Stop recording demonstration and mark as failure \\
Y   & Engage / disengage autonomous policy \\
Joystick & Control base forward and yaw velocity (for door opening) \\
\bottomrule
\end{tabular}
\end{table}

\section{Data Collection Details}
\label{appendix:datacollect}

\begin{figure}[h]
\centering
\begin{subfigure}[b]{.5\textwidth}
  \centering
    \includegraphics[width=\linewidth]{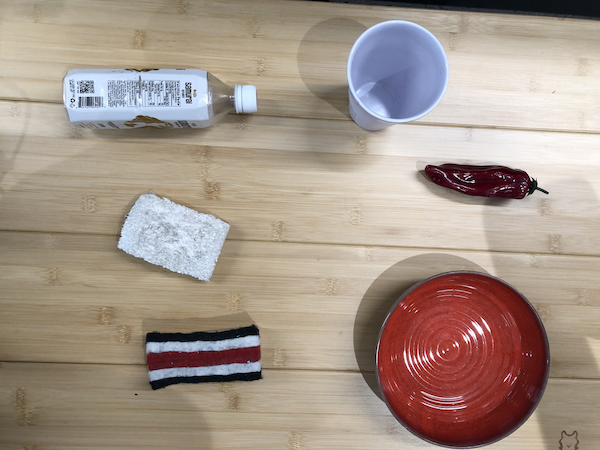}
    \caption{Object Set 1}
    \label{fig:objectset1}
\end{subfigure}%
\begin{subfigure}[b]{.5\textwidth}
  \centering
    \includegraphics[width=\linewidth]{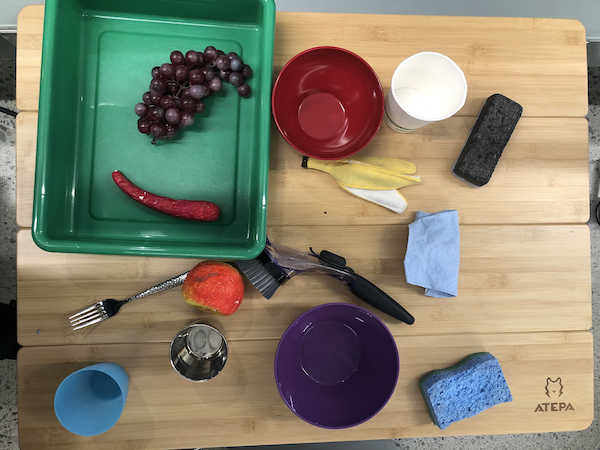}
    \caption{Object Set 2}
    \label{fig:objectset2}
\end{subfigure}%
\caption{Objects used for the 100-task manipulation tasks. Object Set 1 (left) was used to collect data for 21 train tasks, and Object Set 2 (right) was used to collect data for 79 train tasks. For evaluation, 4 tasks were generated between objects in Object Set 2, and \numholdout holdout tasks used objects across both sets. \rebuttal{We used several instances of these objects, with occasional slight differences. For example, the ceramic bowl in Object Set 1 is red in the picture above, but data was also collected with ceramic bowls that were painted green or blue instead.}}
\end{figure}

\rebuttal{\subsection{Inter-task Variability}}

\begin{figure}[h]
\centering
\includegraphics[width=\linewidth]{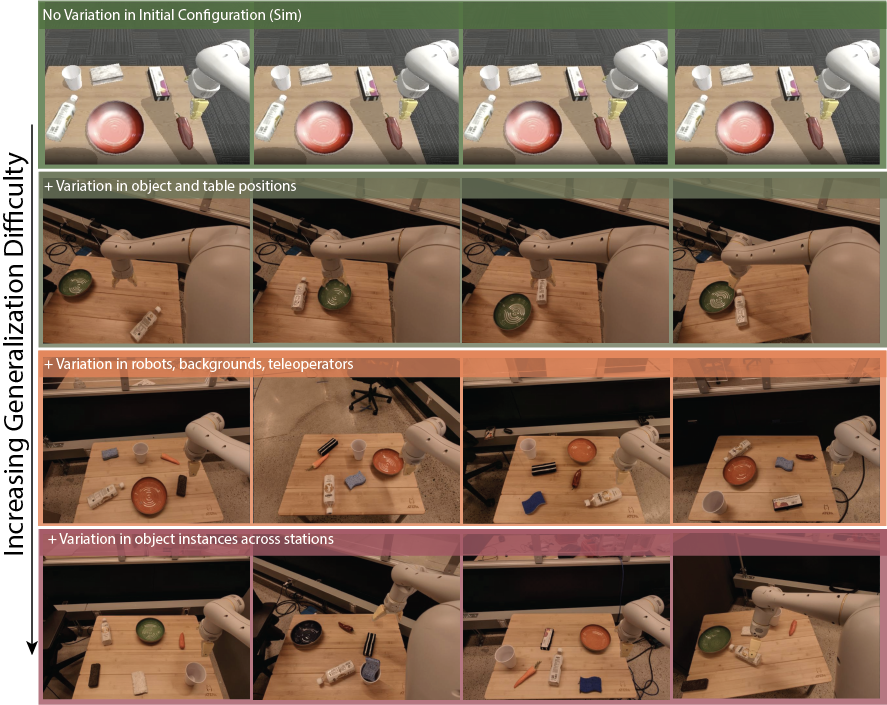}
\caption{\rebuttal{Within each manipulation sub-task (e.g. ``place the bottle in the ceramic bowl'') the policy must generalize not only to variations in object positions, but also varying backgrounds, different robots, and different object instance variations. The top row shows a simulated setup used to confirm that in absence of scene variation, only 37 expert demonstrations are required to solve the task with a 97\% success rate.}} 
\label{fig:subtaskvariability}
\end{figure}

\rebuttal{The data collection protocol is distributed across multiple robots in 1-4 physical locations, resulting in a policy that handles variations across robot hardware, different backgrounds, and scene configurations. Furthermore, each data collection station uses a set of objects with slight physical variations. For instance, the ceramic bowls come in different colors, the sponges can be blue or white with differences in shape, and the erasers and peppers come in different sizes and materials. This variability, illustrated in Figure~\ref{fig:subtaskvariability}, results in a higher sample complexity needed to achieve a desired level of performance on a given set of objects.}

\rebuttal{The low performance of the single-task baseline in Table~\ref{table:ablation-results} trained on 1000 demonstrations begs the question of whether this is due to sample-inefficiency of the behavior cloning implementation, or whether a large number of demos are needed to generalize across the variability in the training data. To study this question, we verify single-task policy performance on ``place the bottle in the ceramic bowl'' in a simulated version of the task, where we can minimize variability across evals and perform exhaustive evaluation of all training checkpoints. When the scene is initialized deterministically with no randomization in initial object positions, 37 expert demos are sufficient to learn the single-task policy with a 97.2\% (0.7) success rate. However, when objects in the scene are randomized, a single-task BC baseline only learns a success rate of 56\% (2.2) when trained on 40 expert demos. These results suggest that the low success rate of the single-task policies in the real setup are indeed caused by the increased diversity in the environment, instead of other factors.}

\subsection{Robotic Teleoperation}

Data for the experiments reported in the paper were collected by \numoperators operators over the course of 5 months. During data collection, tasks are sampled randomly at the beginning of each episode. Objects were occasionally shuffled between episodes, but usually were not, meaning the final state of a demo for one task would often be the start state of a demo for the next task. We found that sampling tasks uniformly was important to performance, since asking teleoperators to pick tasks themselves biased task sampling towards demos that would be easy to perform, creating spurious correlations between initial scene and task demonstrated.

The camera view is kept fixed over the episode. Automatically moving the head to keep the gripper centered in camera frame affords a larger workspace for performing manipulation tasks, but made learning substantially harder.

We experimented with injecting noise into expert actions, following methods like DART~\cite{laskey2017dart}, but found that this compromised the ease of providing demonstrations.

\subsection{Human Video Data Collection}

Human videos were collected in a variety of home and office environments. In each environment, a copy of the scene was set up, and videos were recorded simultaneously by several webcams from different viewpoints. Each viewpoint is treated as a different example of the task, allowing us to collect several videos at once. In general, we found that human videos could be demonstrated 5x-7x faster than a teleoperated robot, so collecting these videos took much less time than collecting the robot dataset.

\begin{figure}
    \centering
    \includegraphics[width=0.45\linewidth]{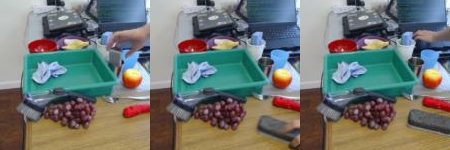}
    \includegraphics[width=0.45\linewidth]{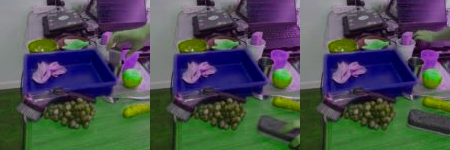}
    \caption{Human demonstrations of the task (left) are augmented with random distortions and reflections (right), then trained to match language features for the task. Robot videos from expert demos are embedded with the same network, along with an end-to-end policy loss.}
    \label{fig:human_demos}
\end{figure}


\section{Featurization Details}
\label{appendix:featurization}

In this work, actions are defined as the difference between states. We found using directly adjacent states ($N = 1$ apart) led to poor performance. Increasing $N$ made action magnitude larger, making the action easier to learn, but introduces bias in the robot trajectory depending on $N$. Through experiments, we found this bias was most problematic when the robot was changing its gripper between open and closed. We therefore used an \textit{adaptive} algorithm to choose $N$. 
For each state, we initialize $N = 1$, then increment it until the change in gripper value of the $N$'th future state exceeds 0.01 or the L2 norm of the arm joint deltas exceeds 0.05. Intuitively, this speeds up the arm when moving the tool across the workspace (higher bias) and slows it down when it is about to come into or out of contact. We also discard labels in which neither the end effector or the gripper are moving.

\section{Policy Training Details}
\label{appendix:training}


A common approach in behavior cloning is to use a Gaussian policy $\pi(a|s,z) = \mathcal{N}(\mu, \sigma)$ to maximize log likelihood of expert actions. We found using a deterministic policy was sufficient to achieve generalization. The network $\pi(a|s,z)$ predicts the delta XYZ, delta axis-angle, and gripper angle (0 to 1). A Huber loss ($\delta$=1.0)~\cite{huber1992robust} is used for XYZ and axis angle, and a log loss is used for gripper angle.

We scaled the XYZ delta losses, axis-angle delta losses, and gripper angle losses by weights of 100, 10, and 0.5, respectively, in order to keep the losses of comparable magnitude for each part of the action.

We implement the model using the FiLM-conditioned ResNet implementation from the open-source Tensor2Robot framework\footnote{\url{https://github.com/google-research/tensor2robot/blob/master/layers/film_resnet_model.py} }. All models are trained using the Adam optimizer with default TensorFlow momentum parameters. When conditioning the model,
we add $\mathcal{N}(0, 0.1)$ Gaussian noise to embedding of the task command,
which was critical to getting the model to predict actions based on the task embedding (as opposed to spurious correlations in the camera image). Multitask manipulation policies were trained with a batch size of 4096 on a TPUv3 pod with a learning rate of $\expnumber{5}{-3}$.

For video task commands,
we found policy performance was stronger if we averaged several video embeddings prior to conditioning.
When evaluating a novel task, we average the embeddings of 10 new human videos collected for the task.

Bin emptying and door opening policies were trained with asynchronous SGD over 10 GPUs with a batch size of 32 per worker and a learning rate of $\expnumber{2.5}{-4}$. In these tasks, we do not add FiLM layers to condition the model, since there is no task command.

\section{Video Conditioning Details}
\label{appendix:video}

For video conditioning, we initially applied a method similar to Task-Embedded Control~\cite{james2018task}. A contrastive loss (cosine similarity or InfoNCE~\cite{oord2018representation}) was applied to the videos, along with the end-to-end BC loss. Embedding visualizations (see Figure~\ref{fig:task_distance}) revealed that the embeddings learned were subpar, collapsing to two main clusters, one for the 21-task family and one for the 79-task family. We hypothesize that because the 21-task family and 79-task family each used a distinct set of objects,
a purely unsupervised objective learned to identify the set of visible objects first, rather than the task performed. Adding the language regression loss helped align the videos more semantically.

\subsection{Video Preprocessing and Architecture}

During training, videos $w$ are randomly subsampled to be 20 frames long, done such that the first and last frame of the video remain in the subsampled $w$. The same image augmentations used for training the policy are applied to the video, but we additionally apply random reflections along the x-axis and y-axis. The augmentation is sampled once per video and applied identically to all 20 frames. At inference time, no augmentations are applied, and the subsampling of 20 frames is done at a uniform frame rate.

The 20 frames are arranged into a 4x5 array of images that are processed by a 2D ResNet-18 network. Arranging the images in this manner allows the 2D convolutions in the ResNet to perform several layers of temporal convolution without needing to tune a new architecture. This was more memory efficient and more performant than a previously tried temporal convolution architecture.
After mean-pooling the final visual features, we add a fully-connected layer with 32 units and ReLU non-linearity, then a linear layer with 512 units and no non-linearity, to match the final size of the language output. The final embedding is normalized to have unit L2 norm. The intermediate 32 unit layer is used to restrict the expressivity of the embedding.

To implement the end-to-end behavior cloning loss, every batch of data must be made of \textit{paired} examples: one human video $w_h \in \mathcal{D}_h^i$ to generate embedding $z$, and one expert demo $\{(s_t,a_t)_{t=1}^T\} \in \mathcal{D}_e^i$ of matching task to act as labels for $\pi(a|s,z)$. We first sample a batch of tasks, with replacement. For each task, we sample 1 human video and 1 robot demo, combining them into one overall batch for the model.
The model is trained using 18 V100 GPUs with a batch size of 28 per GPU. Since each example is 1 pair of videos, every GPU effectively trains on 56 videos at a time. The model was trained with async gradient descent, with a learning rate picked via random search ($\expnumber{2.45}{-4}$).

At train time, the sequence of images $s_t$ in the robot demo are treated as a video of the task, and preprocessed in the same way as the human video. Human videos and robot videos are encoded with the same $q$, and both are trained to match the language embedding with a cosine loss, define as $D_{cos}(v_1, v_2) = 1 - v_1 \cdot v_2$. Embedding noise is not added to $z_h^i$ for the end-to-end loss.

\begin{algorithm}[h]
\SetAlgoLined
\KwInput{Task commands $\mathcal{W}$, per-task robot dataset $\D_e^i$, per-task human video data $\D_h^i$, language encoder $q(\cdot | w_\ell^i)$, video encoder $q(\cdot | w_h)$}
\While{not done training}{
    Sample a batch of tasks $i$, with replacement. \\
    \For{each task $i \in$ batch}{
        Sample human video $w_h \in \D_h^i$ \\
        Sample robot demo $\{(s_t,a_t)\}_{t=1}^T \in \D_e^i$ \\
        Retrieve language command $w_\ell^i$ \\
        $z_h^i \sim q(\cdot | w_h)$  \quad // embed human video \\
        $z_e^i \sim q(\cdot | \{s_t\}_{t=1}^T)$ \quad // embed robot video \\
        $z_\ell^i \sim q(\cdot | w_\ell^i)$ \quad // get language vector \\
        Sample $t \in 1, \cdots, T$ \\
        Compute action $\pi(\hat{a}|s_t,z_h^i)$ \\
        BC-loss $\gets 100 \cdot Huber(xyz) + 10 \cdot Huber(angle) + 0.5 \cdot LogLoss(gripper)$ \\
        Minimize $\mathcal{L} \gets \text{BC-loss} + D_{cos}(z_h^i, z_\ell^i) + D_{cos}(z_e^i, z_\ell^i)$
    }
}
\caption{Pseudocode for training the video encoder}
\label{algorithm:video-encoder}
\end{algorithm}

\subsection{Ablation on Video Encoder Batch}

The sampling strategy for the video encoder batch is non-standard. By sampling tasks, then 1 human video and 1 robot video per task, every task will appear equally often, and every batch is guaranteed to be 50\% human 50\% robot. To examine how this affected model performance, we ran ablations where we first removed the end-to-end behavioral cloning loss, leaving just the language regression objective. Controlling for the same architecture, dataset, hyperparameters, and training time, we change the batch sampling strategy, to either directly sample human and robot videos from all tasks (maintaining a 50-50 batch), or sampling the entire batch uniformly at random over all videos.

Ablations in Table~\ref{table:video-batch} indicate the task-based sampling scheme gives best performance. We hypothesize this sampling does better because it implicitly balances data across both tasks and modalities. The ResNet-18 model also includes batch norm, so it is possible the task-based batch construction affects batch norm in a positive way.

\begin{table}[h]
\caption{Ablations of video encoder batch composition. In the ablations below, we control for the same architecture, dataset, hyperparameters, and training time, changing only the sampling strategy for each batch. The end-to-end behavioral cloning loss is removed, leaving just the language regression loss. Accuracy is measured over held-out videos of training tasks, by checking whether the video embedding is closest to the true language embedding from all 100 train tasks.}
\begin{tabular}{p{9.5cm} p{3.5cm}}
\toprule
    \textbf{Video Encoder Batch Makeup} & \textbf{Video Accuracy} \\
\midrule
    Baseline (sample 28 tasks, pick 1 human + 1 robot video per task) & 84\% \\
    Sample 28 human videos + 28 robot videos & 80\% \\
    Sample 56 videos from entire dataset & 74\% \\
\bottomrule
\end{tabular}
\label{table:video-batch}
\end{table}

\begin{algorithm}[h]
\small
\KwInput{Task commands $\mathcal{W}$, per-task robot dataset $\D_e^i$, per-task human video data $\D_h^i$, language encoder $q(\cdot | w_\ell^i)$, video encoder $q(\cdot | w_h)$, number videos to average $N$}
// Get task vectors, computed once at start of training \\
$taskToVec \gets \{\}$ \\

\For{every task $i$}{
    \eIf{video-conditioned}{
        $z \gets 0$ \\
        \For{$c \gets 1$ \KwTo $N$}{
            Sample video $w_h$ from $\D_{video}$ matching task $i$ \\
            $z_h^i \sim q(\cdot | w_h)$ \\
            $z \gets z + z_h^i$
        }
        $taskToVec[i] \gets L2Normalize(z)$
    }{
        Retrieve language $w_\ell^i$ for task $i$ \\
        $z \gets q(\cdot | w_\ell^i)$ \\
        $taskToVec[i] \gets L2Normalize(z)$
    }
}
// Train policy \\
\For{every epoch}{
    $(i, (s_t, a_t)) \sim \D$ \\
    $z^i \gets taskToVec[i] + \mathcal{N}(0, 0.1)$ \\
    $\loss = 100 \cdot Huber(xyz) + 10 \cdot Huber(angle) + 0.5 \cdot LogLoss(gripper)$ \\
    Update $\pi(a|s,z^i)$ to minimize $\loss$
}
\caption{Pseudocode for training \methodname}
\label{algorithm:bc0}
\end{algorithm}

\section{Video Embedding Visualization}

\begin{figure}
    \centering
    \begin{subfigure}[b]{0.45\textwidth}
        \centering
        \includegraphics[width=\textwidth]{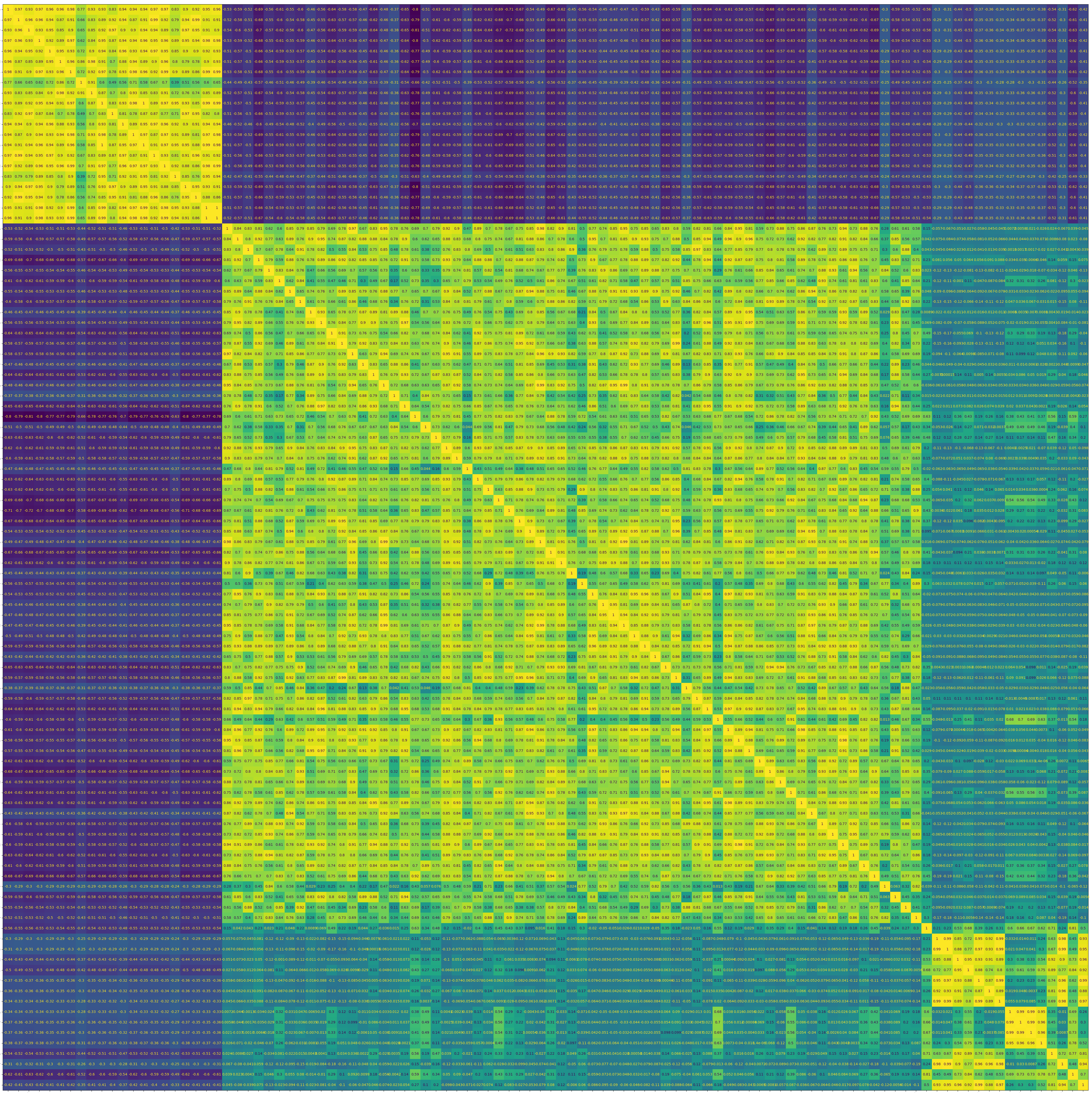}
        \caption{Contrastive baseline}
    \end{subfigure}
    \begin{subfigure}[b]{0.45\textwidth}
        \centering
        \includegraphics[width=\textwidth]{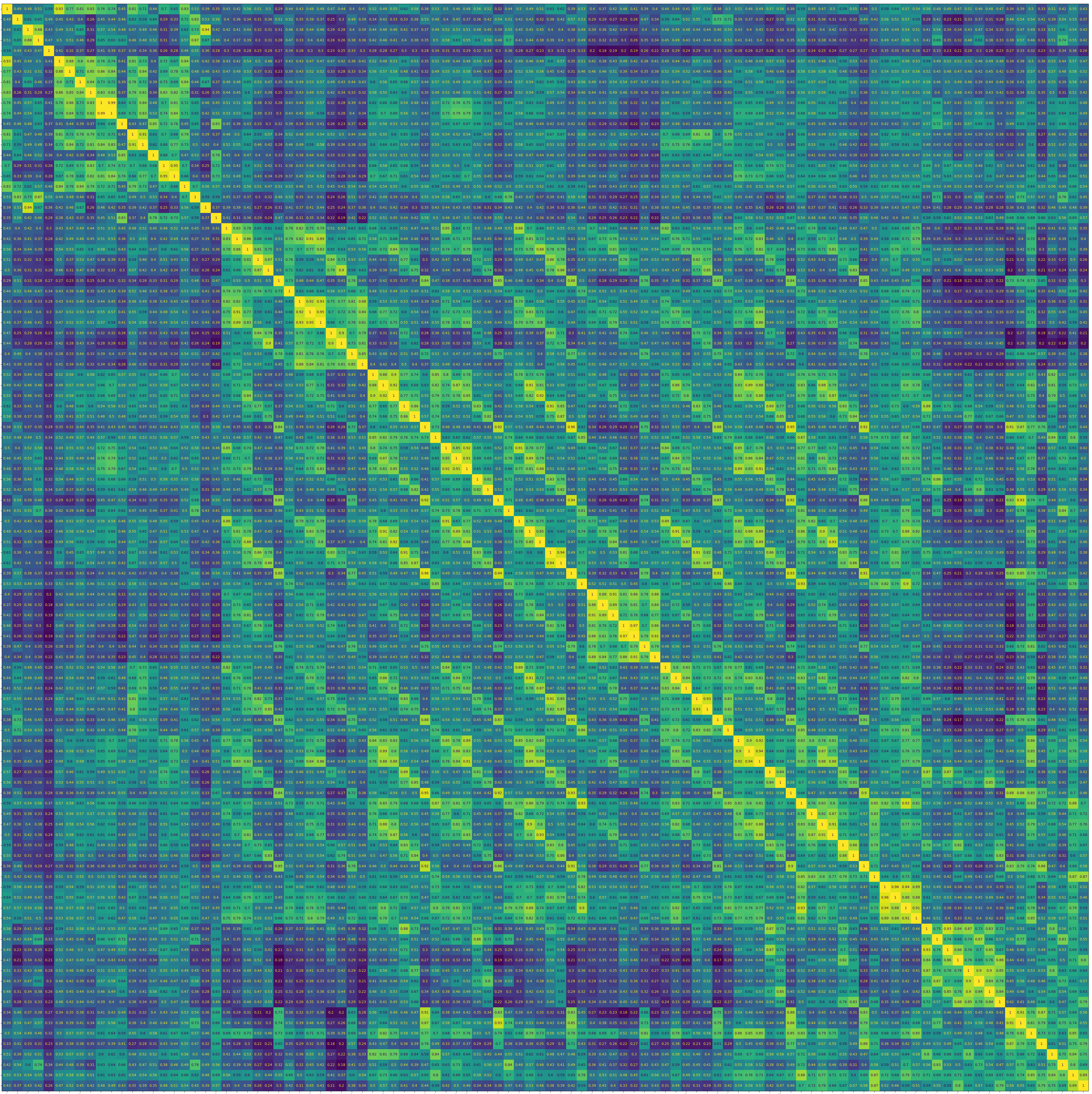}
        \caption{Auxiliary language loss}
    \end{subfigure}
    \caption{Visualizations of different video encoders. Each row and column indicates a different task, with the entry at $(i,j)$ indicating the cosine similarity between video embeddings for task $i$ and task $j$. The 21-task family are the first 21 rows/columns of the visualization. The contrastive baseline learns to primarily group by objects in the scene (by task family), rather than task performed.}
    \label{fig:task_distance}
\end{figure}

Figure~\ref{fig:task_distance} is a similarity matrix across different task embeddings. For each task, we first average the embeddings for all holdout human videos for that task. Entry $(i,j)$ of the matrix is then the cosine similarity between that mean embedding for task $i$ and task $j$. The first 21 rows/columns of the visualization are the 21-task family, and the remaining entries are sorted alphabetically, which groups tasks with the same leading verbs together (i.e. all "place grapes in X" are adjacent to one another.) Adding a language loss term helps prevent the model from grouping tasks based on task family rather than semantic meaning.

\section{Data Annotation Visualization}
\label{appendix:reannotation}

\begin{figure}[h]
  \centering
  \includegraphics[width=\linewidth]{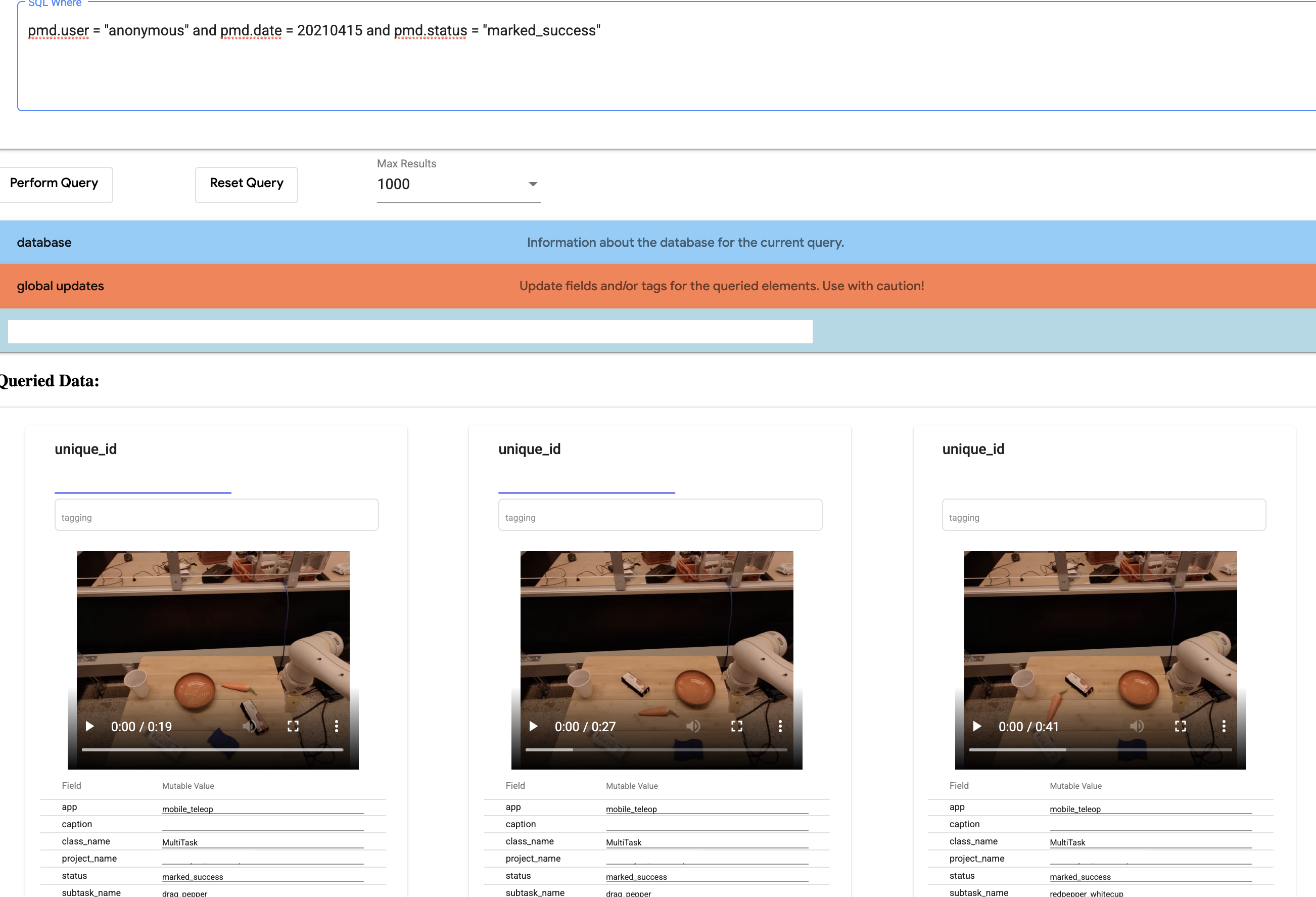}
  \caption{A data browsing and reannotation web interface is used to manually curate episodes and check for low-quality demonstrations. Episodes are searchable via SQL, and their metadata can be edited in-place through the web interface.}
  \label{fig:reannotation}
\end{figure}

During data collection, teleoperators would occasionally record an unsuccessful demonstration as a success, or vice versa. To fix these errors, we built a data visualizer where demos could be retrieved via SQL queries over their metadata, then reannotated. In addition to fixing incorrect labels, we used this tool to perform general data cleaning, such as flagging demonstrations where the robot hardware was faulty, or the arm occluded a target object for the entire demo. The interface is shown in Figure~\ref{fig:reannotation}.

\section{Single-Task Validation on Bin-Emptying}
\label{appendix:sorty-toy-results}

\begin{figure}[t]
  \centering
    \includegraphics[width=\linewidth]{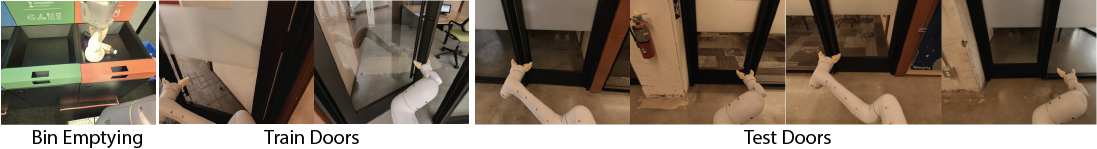}
    \caption{\small Bin Emptying and Door Opening tasks are used to validate that \methodname can achieve a high level of task success and generalize to unseen scenes in a single-task imitation learning setting.}
    \label{fig:singletasks}
    \vspace{-10pt}
\end{figure}

\begin{figure}[h]
\centering
\centering
    \includegraphics[width=.6\linewidth]{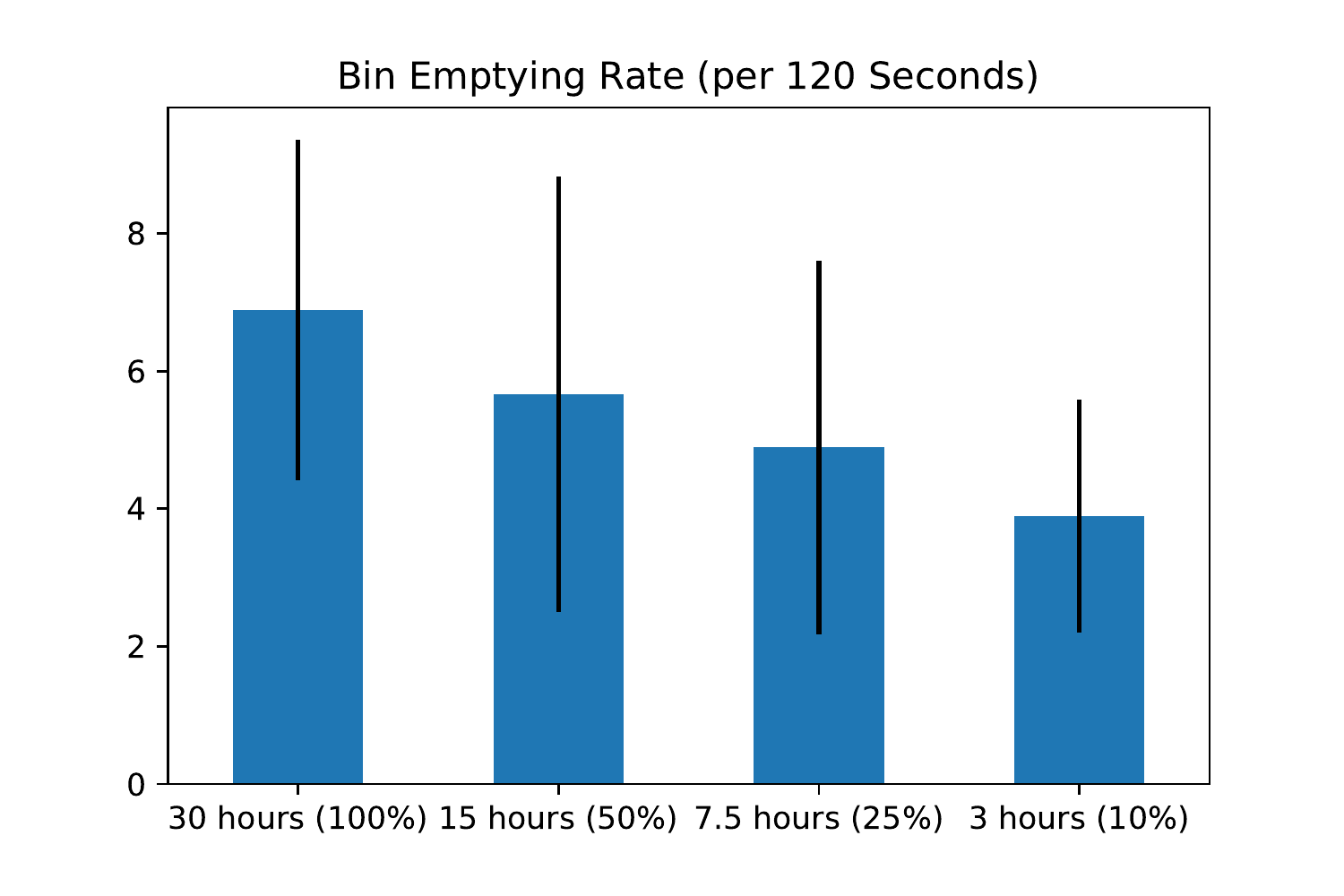}
    \caption{Objects picked in 120 seconds vs. dataset size for bin emptying task}
    \label{fig:bin-empty-rate}
\end{figure}

Figure~\ref{fig:singletasks} illustrates the bin emptying and door opening environments.
In Figure~\ref{fig:bin-empty-rate}, we plot the bin emptying rate as a function of the amount of data used to train the policy. The policy trained on 30 hours of expert demonstrations can clear 3-4 objects a minute. By comparison, a human teleoperator requires about 43 seconds to demonstrate the task. 

An interesting observation to note here is that the bin-emptying task is inherently multimodal, as the objects are grasped in any order during teleoperator demonstrations. The policy learned is a deterministic unimodal policy, and should in principle struggle to learn this multi-model task.
but our model architecture is still able to solve the task.
One hypothesis for why this worked in practice is that the noise and variety within the dataset enabled the model to break symmetry. For instance, the model may have learned to servo towards the nearest object in cases of ambiguity, or it may have used spurious correlations in the background to commit to a specific object.
Whatever the mechanism, it suggests that simple architectures can be sufficient to learn complex tasks, as long as they are trained with appropriate data.

\section{Single-Task Validation on Door Opening}
\label{appendix:door-results}

When validating \methodname on a door opening task, the policy is trained using
12,000 demonstrations collected across 24 meeting rooms, and 36,000 demonstrations across 5 meeting rooms in simulation. For the 24 real meeting rooms, 10 of them swing open from the right, and 14 swing open from the left. For the 5 sim meeting rooms, 3 are right swing, and 2 are left swing.

For each of these demos, the initial arm configuration, as well as
the robot pose with respect to the meeting room's door,
were randomized at the start of each run. Instead of controlling the arm, the policy predicts the forward and yaw velocity of the base. This is an easier control problem, but the policy must still determine the position of the randomized arm relative to the door. The policy also predicts a binary termination action to decide when to stop executing actions, which is predicted with a log loss.
Real-world demonstrations were collected by 13 operators using a fleet of 30 robots. Sim demonstrations were collected using the same interface, with a virtual robot and door rendered on the computer monitor.


To utilize the simulated door opening demonstrations, we used \textit{sim-to-real} transfer to lower the number of required real world demonstrations.
A RetinaGAN model~\cite{retinagan2021} adapted simulated images to look like real images. The RetinaGAN model was trained separately on both sim and real door opening images in an unsupervised fashion by enforcing object-detection consistency between the two domains. Using the RetinaGAN model, we created a third dataset which consists of adapted sim images. We trained the final policy on a mixture of three datasets: sim, real, and adapted sim2real. 

\begin{figure}[t]
  \centering
     \includegraphics[width=0.5\linewidth]{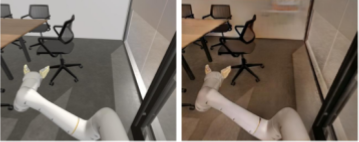}
     \caption{An example of adapting a sim image (left) to look real (right) using RetinaGAN \cite{retinagan2021}.}
  \label{fig:retinagan}
\end{figure}

Policies were evaluated over 4 holdout meeting rooms, made of 2 right swing doors and 2 left swing doors.
For consistency during the evaluation, the operator placed the robot in a certain marked location in front of the meeting room. Then the robot drives to a random pose sampled in an area of 25x25cm from the marked location, and afterwards the trained policy takes control of the robot and performs the task. An episode is marked success if the door is sufficiently open for the robot to enter the room and at the time the policy terminates the task the robot is not in contact with the environment (including the door). Further, any collision of the robot base and arm (not including the gripper) with the environment counted as the task failure by the operator.


\section{Multi-Task Manipulation Training Tasks}
\label{appendix:subtask-results}

Table~\ref{table:traintasks} lists full results of the different task conditioning ablation from Table~\ref{tbl:traintest}, over the 21-task family. The other train tasks in the 79-task family are in Table~\ref{table:remainingtrainsubtasks}.

\begin{table}
\small
  \caption{\small Performance comparsion one-hot, language, or video conditioning over 21 training tasks. Video policies are conditioned on held-out videos of the training tasks. Tasks are ordered by increasing average ``Demo Length'', the number of states observed per expert demonstration, or roughly how many actions the learning policy must learn per episode (both executing at 10 Hz). Length is treated as an estimate of difficulty.
  A subset of eight tasks denoted by \eighttask~are used for comparing additional ablations. Numbers in (parentheses) are 1 unit standard deviation. \emph{See Table~\ref{table:remainingtrainsubtasks} for remaining training tasks.}}
  \label{table:traintasks}
\begin{center}
\resizebox{\columnwidth}{!}{
\begin{tabular}{lp{2cm}lll}
\toprule
\textbf{Tasks From 21-Task Family} & \textbf{Demo Length} & \textbf{One-Hot} & \textbf{Language} & \textbf{Video} \\
\midrule
`knock the eraser over' & 54 (20) & 65\% (7.0) & 91\% (8.7) & 40\% (21.9) \\
`knock the bottle over' & 61 (22) & 58\% (7.4) & 71\% (12.1) & 83\% (15.2) \\
`pick up the ceramic cup'\eighttask & 65 (25) & 67\% (4.7) & 56\% (7.2) &  83\% (15.2) \\
`pick up the ceramic bowl' & 73 (32) & 65\% (7.3) & 77\% (11.7) & 35\% (12.8) \\
`push the ceramic bowl across the table' & 89 (42) & 67\% (8.2) & 58\% (14.2) & 50\% (35.4) \\
`place the pepper in the ceramic bowl' & 98 (30) & 33\% (6.5) & 39\% (11.5) & 12.5\% (11.7) \\
`place the ceramic cup in the ceramic bowl' & 103 (26) & 22\% (6.5) & 33\% (13.6) & 66\% (27.2) \\
`place the white sponge in the ceramic bowl'\eighttask & 106 (32) & 43\% (5.1) & 38\% (6.7) & 14\% (13.2) \\
`place the bottle in the ceramic bowl'\eighttask & 110 (32) & 45\% (5.3) & 52\% (6.3) & 43\% (18.7) \\
`drag the pepper across the table'\eighttask & 115 (43) & 55\% (5.0) & 33\% (8.2) & 20\% (12.6) \\
`push the eraser across the table'\eighttask & 117 (50) & 71\% (4.7) & 45\% (7.9) & 0\% (0) \\
`place the eraser on the white sponge'\eighttask & 121 (37) & 33\% (4.9) & 30\% (7.2) & 25\% (15.3) \\
`place the pepper on the white sponge' & 123 (44) & 36\% (7.1) & 83\% (10.8) & 0\% (0) \\
`place the pepper in the ceramic cup'\eighttask & 128 (44) & 30\% (4.6) & 49\% (7.6) & 0\% (0) \\
`place the ceramic cup over the eraser' & 130 (38)  & 9\% (5.0) & 0\% (0) & 0\% (0) \\
`place the white sponge in the ceramic cup' & 131 (38) & 21\% (6.9) & 25\% (10.8) & 14\% (13.2) \\
`place the eraser in the ceramic cup'\eighttask & 135 (43) & 37\% (4.8) & 33\% (7.5) & 20\% (17.9) \\
`move the arm in a circular motion' & 144 (69) & 69\% (6.7) & 12\% (6.8) & 0\% (0) \\
`drag the ceramic bowl in a circle' & 164 (65) & 32\% (6.8) & 0\% (0) & 0\% (0) \\
`wipe the white sponge on the table' & 177 (59) & 29\% (7.8) & 18\% (9.2) & 0\% (0) \\
`stand the bottle upright' & 186 (55) & 5\% (3.5) & 0\% (0) & 0\% (0) \\
\midrule
\textit{Overall} & 115 & 42\%& 40\% & 24\%  \\
\bottomrule
\end{tabular}
}
\end{center}
\end{table}

\begin{table}
  \caption{Additional multitask manipulation training task sentences}
  \label{table:remainingtrainsubtasks}
\begin{tabular}{l}
\toprule
\textbf{Name} \\
\midrule
'wipe purple bowl with sponge' \\
`wipe red bowl with sponge' \\
`wipe table surface with sponge' \\
`wipe purple bowl with towel' \\
`wipe red bowl with towel' \\
`wipe tray with towel' \\
`wipe table surface with towel' \\
`wipe purple bowl with brush' \\
`wipe red bowl with brush' \\
`wipe tray with brush' \\
`wipe table surface with brush' \\
`wipe purple bowl with eraser' \\
`wipe red bowl with eraser' \\
`wipe tray with eraser' \\
`wipe table surface with eraser' \\
`place sponge in purple bowl' \\
`place sponge in red bowl' \\
`place sponge in table surface' \\
`place sponge in metal cup' \\
`place sponge in plastic cup' \\
`place sponge in paper cup' \\
`place towel in purple bowl' \\
`place towel in red bowl' \\
`place towel in tray' \\
`place towel in table surface' \\
`place towel in metal cup' \\
`place towel in plastic cup' \\
`place towel in paper cup' \\
`place brush in purple bowl' \\
`place brush in red bowl' \\
`place brush in tray' \\
`place brush in table surface' \\
`place brush in metal cup' \\
`place brush in plastic cup' \\
`place brush in paper cup' \\
`place eraser in purple bowl' \\
`place eraser in red bowl' \\
`place eraser in tray' \\
`place eraser in table surface' \\
`place eraser in metal cup' \\
\end{tabular}
\quad
\begin{tabular}{l}
\toprule
\textbf{Name}  \\
\midrule
`place eraser in plastic cup' \\
`place eraser in paper cup' \\
`place grapes in purple bowl' \\
`place grapes in tray' \\
`place grapes in table surface' \\
`place grapes in metal cup' \\
`place grapes in plastic cup' \\
`place grapes in paper cup' \\
`place banana in purple bowl' \\
`place banana in red bowl' \\
`place banana in tray' \\
`place banana in table surface' \\
`place banana in metal cup' \\
`place banana in plastic cup' \\
`place banana in paper cup' \\
`place apple in purple bowl' \\
`place apple in red bowl' \\
`place apple in tray' \\
`place apple in table surface' \\
`place apple in metal cup' \\
`place apple in plastic cup' \\
`place pepper in purple bowl' \\
`place pepper in red bowl' \\
`place pepper in tray' \\
`place pepper in table surface' \\
`place pepper in metal cup' \\
`place pepper in plastic cup' \\
`place pepper in paper cup' \\
`place fork in purple bowl' \\
`place fork in red bowl' \\
`place fork in tray' \\
`place fork in table surface' \\
`place fork in metal cup' \\
`place fork in plastic cup' \\
`place fork in paper cup' \\
`stack cups on top of each other' \\
`stack bowls on top of each other' \\
`stack cups into tray' \\
`stack bowls into tray' \\
\\ 
\end{tabular}
\end{table}



\section{Details on HG-DAgger Ablation}
\label{app:ablationdetails}

Two models were trained, one using all expert-only demonstrations, and one using a 50-50 mixture of expert-only demonstrations and HG-DAgger demonstrations. The 50-50 dataset has the same number of episodes as the expert-only dataset. To reduce evaluation time, the HG-DAgger comparison was evaluated on a smaller subset of 8 tasks from the 21-task family, indicated in Table~\ref{table:traintasks} by the \eighttask~symbol.





%

\section{Negative Results}
\label{appendix:negative-results}

Below is an incomplete list of alternative methods tried during the course of the research project. These are empirical observations and are presented as-is. These ideas were generally tried once or twice, then set aside when they did not improve performance. We did not do a detailed investigation of the negative results, so it is possible they did not perform better due to incorrect implementation, the presence of a different performance bottleneck in the system, or improperly tuned hyperparameters, rather than deficiencies in their ideas. We hope this anecdotal experience will be helpful to researchers building on top of this work.


\begin{itemize}
    \item The policy is trained with a Huber loss ($\delta=1.0)$), that in practice is usually just an MSE loss, since predictions almost always lie within $(-1, 1)$. A deterministic policy trained with mean-squared error can be viewed as max likelihood with a unimodal Gaussian policy, with learned mean $\mu$ and fixed $\sigma$. We tried a stochastic policy, with a learned $\sigma$ based on the current state, but found it did not help and seemed to make training less stable.
    \item Similarly, a mixture density network (mixture of 10 Gaussians) did not improve performance.
    \item Using a larger model architecture than ResNet-18 (ResNet-34 and larger) also did not improve performance.
    \item To address the small action problem identified in Section~\ref{section:policy-training}, we tried decomposing XYZ prediction into direction and magnitude. The hypothesis was that by making it easier for the model to predict small actions, it would prevent the model from predicting small actions at every state. This did not outperform predicting XYZ directly, and eventually led to the adaptive algorithm used in the final results.
    \item We initially used a spatial softmax layer in our policy and video encoder. Visualizing those spatial softmax layers made it easier to interpret policy predictions, but performance increased when the spatial softmaxes were removed.
    \item Conditioning the policy on proprioceptive information, as well as previous robot poses, did not improve performance. It is possible this was due to causal confusion between that information and the expert actions~\cite{de2019causal}.
    \item Using more video frames (40 instead of 20) did not improve performance of the video encoder, and slowed down training.
    \item We experimented with including human videos that did not correspond to any of the robot tasks, using them as negative examples for a contrastive loss, to encourage the task embeddings to be more continuous. We found this did not help, and the negative examples were too easy to embed far away from all other videos.
    \item Pre-training the ResNet on the ILSVRC2012 object classification dataset did not improve performance.
    \item We obtained better results on manipulation tasks by representing angles as delta axis-angle, rather than absolute axis-angle, absolute quaternions, or delta quaternions.
    \item At each state, the policy predicts a 10 action long open-loop trajectory, then only executes the first action. Stopping the gradient from the 2nd to 10th predicted actions of the open-loop trajectory was inconclusive but generally did not help. 
    \item Applying mixup regularization~\cite{zhang2018mixup} to the images and robot poses did not help. We suspected it played poorly with the continuous outputs, and might work better if actions were discretized.
    \item Predicting gripper residuals instead of the absolute open/close angle.
    \item We found that while validation error on predicting future poses was correlated with task success, different models with similar validation errors could have wildly varying levels of task success. This made selecting the right checkpoint for evaluation challenging; it is quite possible that performance numbers would be higher with additional evaluation budget for specific checkpoints. This is likely because validation accuracy is most critical on specific states (especially near contact), while there is more tolerance for error on other states.
\end{itemize}

\rebuttal{\section{Further Video Embedding Comparisons}}

We perform further experiments on alternative video embedding methods. Each method is trained on the same HG-DAgger dataset that was collected using the \methodname system and policies.
We consider two alternative embedding methods. First, since the pre-trained language embeddings showed considerable success, we consider a pre-trained video embedding~\cite{miech2020end} that was trained on a large set of instructional videos.
Second, we also compare to the embedding approach introduced by~\citep{james2018task}, referred to as TecNets, which trains the embedding using the end-to-end policy objective and a contrastive loss between different videos (with no language component). In this comparison, we retrain the policy on top of the frozen embedding after it is learned, as in \methodname, as we find this to improve training stability.

The results are shown in Table~\ref{table:rebuttaltasks}.
Pre-trained video embeddings enable some generalization to held-out tasks, but do not perform quite as well as the learned video embedding in \methodname. In our visualizations, we find that pretrained video embeddings are quite similar across different tasks, which may have hurt multitask learning from those embeddings. We theorize that these pre-trained models tend to group videos more based on the background scene than based on actions taken. In the TecNet comparison, we find that the TecNets video embedding leads to worse overall performance on held-out tasks. In particular, the results show that TecNets has similar performance to video-conditioned \methodname on the 4 held-out tasks that only use objects from the 79-task family, but that performance is worse on 3 held-out tasks that mixed objects between the 21-task family and 79-task family. Anecdotally, we also found that the TecNet training runs were less stable: in two runs only differing by random seed, one embedding collapsed and the other did not (Table~\ref{table:rebuttaltasks} reports performance of the training run that did not collapse).

\begin{table}[h]
\small
  \caption{\small Performance comparison between different video embeddings on selected tasks. All tasks are held-out unless otherwise indicated. Numbers in (parentheses) are 1 unit standard deviation}
  \label{table:rebuttaltasks}
\begin{center}
\resizebox{\columnwidth}{!}{
\begin{tabular}{llll}
\toprule
\textbf{Task} & \textbf{Video-conditioned \methodname} & \textbf{\methodname, pretrained video embedding} & \textbf{\methodname, TecNet embedding} \\
\midrule
`place sponge in tray' & 22\% (2.2) & 10\% (6.7) & 0\% (0)  \\
`place grapes in red bowl' & 12\% (7.8) & 30\% (10.2) & 20\% (8.9) \\
`place apple in paper cup' & 14\% (7.8) & 0\% (0)  & 25\% (9.7) \\
`wipe tray with sponge' & 28\% (10.6) & 15\% (8.0) & 15\% (8.0) \\
`place banana in ceramic bowl' & 7.5\% (4.2) & 0\% (0)  & 0\% (0)  \\
`pick up bottle' & 17.5\% (6.0) & 10\% (6.7) & 0\% (0)  \\
`push purple bowl across the table' & 0\% (0) & 0\% (0) & 0\% (0)  \\
\midrule
\textit{Average (held-out)} & 14.4\% & 9.3\% & 8.6\%  \\
\bottomrule
\end{tabular}
}
\end{center}
\end{table}

%

%

\end{document}